\documentclass[]{fairmeta}
\usepackage{wrapfig}
\usepackage{tabularx}
\usepackage{textcomp}
\usepackage{stfloats}
\usepackage{url}
\usepackage{verbatim}
\usepackage{graphicx}
\usepackage{titlesec}
\usepackage{tocloft}
\usepackage{adjustbox}
\usepackage{multirow}
\usepackage{pifont}
\usepackage{tikz}
\usepackage{comment}
\usepackage{amsmath,amssymb} %
\usepackage{colortbl}  %
\usepackage[dvipsnames]{xcolor}         %
\usepackage{booktabs} 
\usepackage{hyperref}
\usepackage{graphicx}    %
\usepackage{subcaption} 
\usepackage{booktabs}
\usepackage{amsmath} %
\usepackage{multirow} %
\usepackage{booktabs} %
\usepackage{subcaption} %
\usepackage{todonotes}
\usepackage{natbib}
\usepackage{makecell}
\usepackage{tcolorbox,enumitem,ragged2e}
\tcbset{
  promptbox/.style={
    colback=black!2, colframe=black!20, boxrule=0.3pt,
    arc=1mm, left=6pt, right=6pt, top=6pt, bottom=6pt}
}

\RequirePackage{xspace}
\makeatletter
\DeclareRobustCommand\onedot{\futurelet\@let@token\@onedot}
\def\@onedot{\ifx\@let@token.\else.\null\fi\xspace}

\makeatother

\definecolor{adptorange}{RGB}{248, 205, 172}
\definecolor{cmpblue}{RGB}{189, 215, 238}
\definecolor{cmpblue}{RGB}{189, 215, 238}

\definecolor{our_red}{RGB}{232,157,160}
\definecolor{our_blue}{RGB}{136,206,230}
\definecolor{our_orange}{RGB}{246,200,168}
\definecolor{our_green}{RGB}{178,211,164}

\definecolor{attn_code0}{RGB}{247,215,200}
\definecolor{attn_code1}{RGB}{238,169,139}
\definecolor{mlp_code0}{RGB}{204,201,221}
\definecolor{mlp_code1}{RGB}{102,95,153}

\definecolor{dark_green}{rgb}{0, 0.5, 0}
\definecolor{dark_red}{rgb}{0.8, 0.2, 0.2}
\definecolor{soft_red}{rgb}{1.0, 0.4, 0.4}
\definecolor{light_blue}{rgb}{0.2, 0.5, 1.0}

\usepackage{algpseudocode}
\usepackage{algorithm}

\definecolor{token_blue}{RGB}{84, 120, 140}

\usepackage{pifont}       %
\usepackage{bbding}       %
\usepackage{fontawesome}
\usepackage{xspace}

\usepackage{float}
\usepackage{siunitx}        %
\usepackage{microtype}      %
\usepackage{algorithm}     %
\usepackage{algorithmicx}  %
\usepackage{algpseudocode} %
\usepackage{cleveref}      %
\usepackage{enumitem}
\usepackage[inkscapelatex=false]{svg}
\usepackage{animate}
\usepackage{etoc}          %

\definecolor{darkgreen}{rgb}{0.15, 0.75, 0.15}
\definecolor{cvprblue}{rgb}{0.21,0.49,0.74}
\definecolor{lightblue}{rgb}{0.90, 0.95, 0.99}

\algrenewcommand\algorithmicrequire{\textbf{Input:}}
\algrenewcommand\algorithmicensure{\textbf{Output:}}

\title{
  WorldAttention: An Efficient Attention Architecture for Interactive Video World Models
}
\author[1*]{Zeyu Zhang}
\author[2*]{Jinyuan Mao}
\author[3*]{Dakai An}
\author[3]{Wangbo Zhao}
\author[3]{Hanfeng Lu}
\author[1,4\dag]{Jiasheng Tang}
\author[5]{Yinghao Yu}
\author[3]{Wei Wang}
\author[1,2\dag]{Bohan Zhuang}

\affiliation[1]{DAMO Academy, Alibaba Group}
\affiliation[2]{Zhejiang University\\}
\affiliation[3]{Hong Kong University of Science and Technology}
\affiliation[4]{Hupan Lab}
\affiliation[5]{TRE, Alibaba Group}
\contribution[*]{Equal contribution}
\contribution[\dag]{Corresponding authors.}

\abstract{
Leveraging the paradigm of autoregressive diffusion, text-conditioned interactive video world models aim to simulate temporally coherent environments guided by textual instructions. While enabling low-latency, long-duration generation is pivotal for embodied AI and simulation-based planning, current frameworks primarily rely on sliding-window mechanisms to bound computational complexity. However, this approach inherently sacrifices historical context, undermining the long-range interactive capabilities. Conversely, maintaining a full-history cache remains computationally prohibitive and memory-intensive: the quadratic complexity of attention leads to excessive computational overhead, while the linear growth of the KV cache inevitably leads to GPU memory saturation.
To overcome these limitations, we propose \textbf{WorldAttention}, a system-oriented attention architecture that achieves high efficiency through the co-design of specialized attention kernels and hierarchical KV cache management.
First, we introduce \textit{Hybrid Sparse Attention (HSA)}, which integrates linear global attention supplemented with head-adaptive sparse attention. Additionally, we design a \textit{Hierarchical KV Cache (HKV)} that organizes historical KV pairs into semantically indexed pages across multi-tier memory, enabling fine-grained retrieval and controlled GPU residency. These two designs are supported by tailored kernels to effectively translate their theoretical efficiency into real-world performance. Extensive experiments on VBench-Long and InterVBench demonstrate that WorldAttention consistently surpasses prior state-of-the-art methods, achieving subject consistency scores of \textbf{0.9472} on VBench-Long and \textbf{0.9668} on InterVBench, respectively. Our system-oriented %
kernel customization for HSA brings a \textbf{14.02$\times$} speedup over FlashAttention-3, and a \textbf{2.21$\times$} end-to-end speed up together with HKV. At inference, WorldAttention sustains 22.0 FPS on a single NVIDIA H100.}

\date{\today} 
\metadata[Website]{\url{https://alibaba-damo-academy.github.io/WorldAttention}}
\metadata[Code]{\url{https://github.com/alibaba-damo-academy/WorldAttention}}
\metadata[Email]{\email{jiasheng.tjs@alibaba-inc.com}, \email{bohan.zhuang@gmail.com}}

\begin{document}
\thispagestyle{firstheader}
\maketitle

\section{Introduction}

Text-conditioned interactive video world models~\cite{zhang2025bife,yang2025longlive} aim to generate dynamic, coherent video environments that respond to user prompts over time. 
A key challenge unique to this setting is \emph{interactive prompt switching}: user inputs can abruptly shift semantic focus and require recalling fine-grained visual details from previous memory. 
This fundamentally distinguishes interactive generation from standard long video modeling, as the model must maintain long-range identity, scene consistency, and causal coherence across the entire history.
As a result, the KV cache must preserve the full history to support interaction recall.
However, this creates a fundamental tension between memory preservation and system efficiency: while preserving long-range memory is necessary for interaction, the KV cache grows linearly with sequence length and quickly exceeds GPU memory capacity, limiting scalability for long-horizon interactive generation.

Existing KV cache strategies make different trade-offs between memory efficiency and recall ability. 
(1) \textit{Sliding-window KV cache}~\cite{cui2026self,huang2025self,yang2025longlive,chen2025skyreels,yi2025deep} retains only recent tokens, which reduces GPU memory usage but permanently discards earlier visual context. 
When a user prompt references a scene from tens of seconds earlier, the required visual details are irretrievably lost, making sliding-window caching functionally inadequate for interactive generation. 
(2) \textit{Sparse retrieval KV cache}~\cite{zhang2025bife}, in contrast, preserves all KV cache to support interactive recall. 
However, retaining the full history immediately introduces severe storage pressure, as the KV cache grows linearly with video length and quickly exceeds GPU memory capacity. 
This creates a hard constraint: full-history KV must be maintained for interaction, but must be efficiently managed, necessitating an efficient attention architecture for both attention re-design and hierarchical caching mechanism.

To address these challenges, we propose \textbf{WorldAttention}, an efficient attention architecture with %
tailored kernels that aims at memory management and computation complexity in interactive video generation, as shown in Figure \ref{fig:worldattention}.
First, to enable full-history restoration under limited GPU memory, we introduce \textit{Hierarchical KV Cache (HKV)}, which implements a fine-grained page-level storage and retrieval mechanism across multi-tier memory (GPU, CPU, and NVMe). This allows the system to maintain a vast historical context while minimizing GPU memory residency.
However, the retrieved KV pages still contain a large number of tokens, leading to significant redundant computation during decoding. 
To address this, we further introduce \textit{Hybrid Sparse Attention (HSA)}, which performs dual-branch sparse attention within retrieved pages via head-adaptive sparsity. 
Finally, hierarchical retrieval and sparse execution introduce non-contiguous memory access patterns, which severely degrade hardware efficiency. 
To resolve this, we %
adapt the \textit{attention kernel} with KV layout, reorganizing retrieved pages into contiguous buffers and enabling efficient parallel execution.
Extensive experiments demonstrate that WorldAttention consistently outperforms prior state-of-the-art methods, achieving subject consistency scores of \textbf{0.9472} on VBench-Long and \textbf{0.9668} on InterVBench, which shows the improvement of our method on  interactive video generation over long horizons, while significantly reducing memory and computation overhead.

Our contributions are summarized as follows:
\begin{itemize}[leftmargin=*]
    \item We propose \textbf{WorldAttention}, an efficient attention architecture specifically designed for interactive long video generation to systematically overcome the constraints of quadratic attention computation and growing memory footprint.
    
    \item We introduce a system-oriented co-design that integrates Hierarchical KV Cache (HKV) for paged storage and retrieval with Hybrid Sparse Attention (HSA) to reduce computation redundancy. Our system-oriented %
    kernel integration for HSA brings a \textbf{14.02$\times$} speedup over FlashAttention-3, and a \textbf{2.21$\times$} end-to-end speed up together with HKV.
    
    \item We demonstrate through comprehensive evaluations that our method achieves state-of-the-art performance, enabling scalable, coherent, and high-fidelity interactive video generation over minute-long horizons.
\end{itemize}

\section{Related Work}

\paragraph{KV Cache in video world models.} In video generation, related formulations adopt chunk-wise diffusion with causal conditioning to interpolate between pure diffusion (e.g. DiT-style bidirectional attention without KV caching) and AR (variable-length decoding with KV caching but weaker visual fidelity and limited parallelism). Representation methods with a \textit{sliding-window KV cache} include MAGI-1~\cite{teng2025magi}, Self Forcing~\cite{huang2025self}, Self-Forcing++~\cite{cui2026self}, Rolling Forcing~\cite{liu2025rolling}, Deep Forcing~\cite{yi2025deep}, StreamDiffusionV2~\cite{feng2025streamdiffusionv2}, PAVDM~\cite{xie2025progressive}, LongLive~\cite{yang2025longlive}, CausVid~\cite{yin2025causvid}, SkyReels-V2~\cite{chen2025skyreels}, and Infinity-RoPE~\cite{yesiltepe2025infinity}, etc, which condition each new chunk on past chunks via a sliding-window KV cache to extend temporal horizons while retaining diffusion denoising quality within each chunk. Another paradigm, the \textit{sparse retrieval KV cache}~\cite{zhang2025bife}, introduces a semantic sparse KV cache that retrieves the most relevant chunk-level prompts, effectively maintaining long-horizon memory during interactive generation.
However, the continuously growing KV cache and the relatively coarse retrieval granularity become new bottlenecks. To overcome these limitations, WorldAttention proposes a hierarchical KV cache management framework with page-level sparse retrieval, further improving efficiency and generation quality.

\paragraph{Sparse and linear attention in video generation.} 

Recent advances in video generation increasingly rely on structured sparsity and linearized attention to improve efficiency. 
\textit{Sparse attention} methods reduce quadratic complexity by pruning token interactions. 
Sliding Tile Attention~\cite{zhang2025fast,liu2025fpsattention} introduces tiled sparse patterns to model spatial-temporal correlations, while SpargeAttn~\cite{zhang2025spargeattention} progressively prunes tokens based on importance scores. 
DiTFastAttn~\cite{yuan2024ditfastattn} dynamically filters irrelevant patches to accelerate computation, and Sparse VideoGen~\cite{xi2025sparse,yang2025sparse} leverages sparsely sampled motion priors to reduce temporal redundancy. 
More recent structured block-sparse designs, such as VSA~\cite{zhang2025vsa}, align sparsity with GPU execution to achieve practical speedups.
\textit{Linear attention} methods approximate global attention with low-rank or kernel-based projections to achieve $\mathcal{O}(L)$ complexity. 
Representative approaches include LinFormer~\cite{wang2020linformer}, Gated Linear Attention~\cite{yang2023gated}, and Tiled Flash Linear Attention~\cite{beck2025tiled}. Moreover, \textit{hybrid sparse-linear formulations} such as SEA~\cite{lee2023sea} and Sparse Linear Attention (SLA)~\cite{zhang2025sla}, which combine coarse global modeling with local sparse refinement.
While these approaches demonstrate the effectiveness of structured sparsity and linearization for scaling video transformers, they are often designed primarily for inference-time acceleration, lack adaptation to chunk-wise long-horizon video generation, and do not explicitly co-design attention structure with KV management and hardware-aware kernel execution.

\section{Method}

\subsection{Overview}

\begin{figure}[t]
    \centering
    \includegraphics[width=\linewidth]{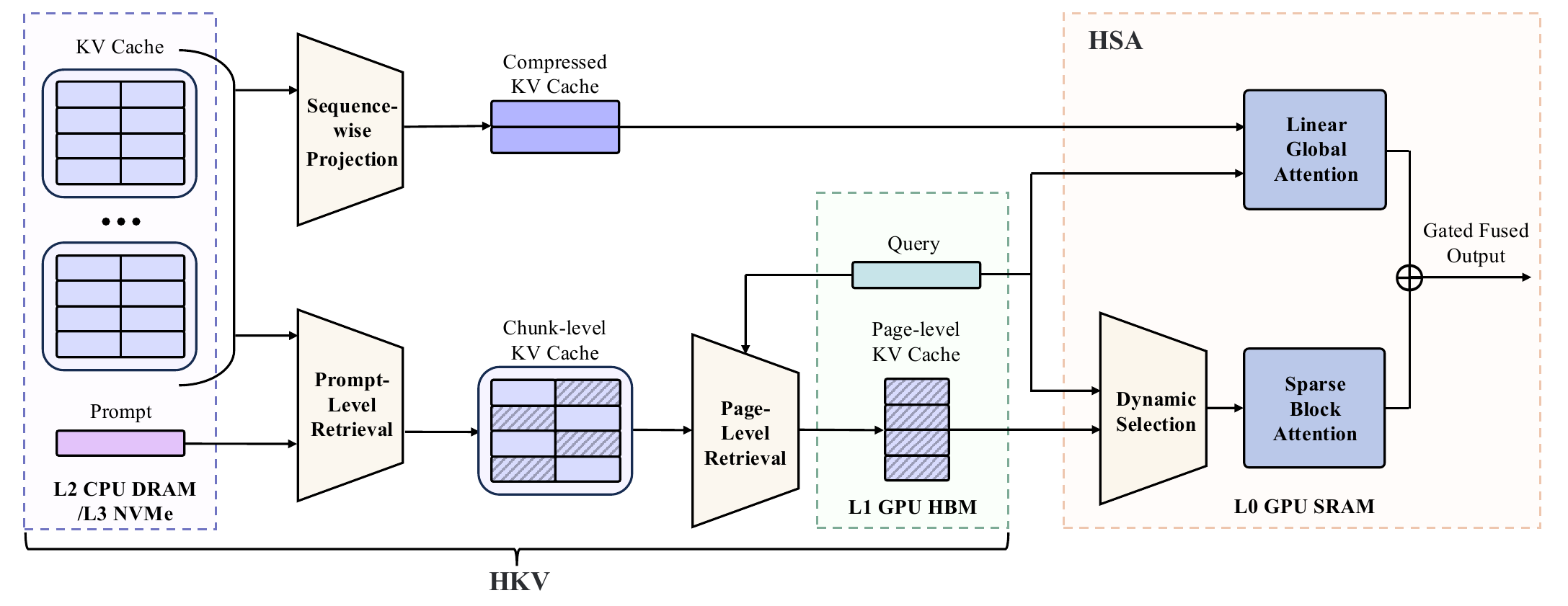}
    \caption{\textbf{Overall architecture of WorldAttention.} We propose a system-oriented co-design using Hierarchical KV Cache (HKV) for coarse-to-fine memory retrieval, and Hybrid Sparse Attention (HSA) for efficient dual-branch attention computation.}
    \label{fig:worldattention}
\end{figure}

Figure~\ref{fig:worldattention} illustrates the overall pipeline of WorldAttention for interactive long video generation. 
Given an initial text prompt, the video is generated chunk by chunk under an autoregressive diffusion framework. 
At each decoding step, the current prompt and latent chunk produce query tokens, while historical chunks provide cached keys and values (KV) for long-range conditioning.
Unlike standard long video generation, interactive generation allows user prompts to change across chunks, requiring the model to retrieve relevant historical context from earlier video segments. 
To support this, we maintain a global KV cache over all past chunks and dynamically select relevant KV entries during decoding.
WorldAttention is the core attention module that operates on the retrieved KV cache. 
It consists of three components: (1) a \textit{Hierarchical KV Cache (HKV)} that organizes and retrieves KV entries across memory tiers, 
(2) a \textit{Hybrid Sparse Attention (HSA)} that performs efficient attention over the retrieved KV tokens, and
(3) a customized kernel integration with memory-layout optimization that enables efficient execution.

\subsection{Hierarchical KV Cache}
\label{sec:hkvc}

KV cache in interactive long video generation faces two significant challenges:
First, existing designs lack cross-tier data migration, forcing all KV cache to remain in GPU memory and causing memory usage to grow with video length. 
Second, retrieval is typically performed at the chunk level, loading entire chunks even though only a small subset of tokens is relevant, leading to redundancy and imprecise retrieval.
These issues stem from a mismatch between memory organization and attention usage: relevant context is often localized, while KV cache is managed at coarse chunk granularity. 

Therefore, we introduce a \textit{page-level KV abstraction}, which partitions each chunk into semantically coherent KV pages $\mathcal{P}$, where each page contains 8 video frames, as shown in Figure~\ref{fig:pipeline}.
This enables fine-grained retrieval and efficient cross-tier memory management, forming the foundation of our \textit{Hierarchical KV Cache (HKV)}.

\paragraph{Cache hierarchy and dataflow.}
The hierarchical cache organizes KV cache into multiple storage tiers:
\textit{L0 (Accelerator-Managed)}: a GPU SRAM serves as the high-speed on-chip memory used for computing Q, K, and V in FlashAttention-Style.
\textit{L1 (Device-Managed)}: a GPU HBM stores the retrieved KV pages from the KV bank, along with chunk indices composed of prompt embeddings and page indices constructed from the keys $K$.
\textit{L2 (Host-Managed)}: a CPU DRAM contains KV chunks that are not retrieved during the generation of the current chunk.
\textit{L3}: an NVMe stores all KV chunks, serving as the final fallback for historical context.
This design aligns access frequency with hardware bandwidth, allowing retrieved KV pages to remain in GPU memory, while less frequently accessed KV chunks are automatically offloaded to the larger and lower-cost CPU memory. This enables more accurate historical guidance for the current chunk generation.

\paragraph{Global KV bank initialization.}

All historical KV pairs produced by HSA are first written into a unified
\emph{KV bank}, denoted as
$
\mathcal{B}
=
\left\{
\mathcal{C}_1, \mathcal{C}_2, \dots, \mathcal{C}_M
\right\},
$
where each $\mathcal{C}_m$ corresponds to a KV chunk generated from a video segment.
Each chunk $\mathcal{C}_m$ is further partitioned into fixed-size KV pages,
\begin{equation}
\mathcal{C}_m
=
\left\{
\mathcal{P}_{m,1}, \dots, \mathcal{P}_{m,J}
\right\},
\qquad
\mathcal{P}_{m,j}
=
\left(
\mathbf{K}_{m,j},\mathbf{V}_{m,j}
\right),  
\end{equation}
where each page contains KV pairs of 8 consecutive frames.

\paragraph{Stage 1: prompt-level retrieval.}

When generating the current chunk $t$, we first retrieve the top-1 KV chunk by computing cosine similarity between the current prompt embedding and previous-chunk prompt embeddings in the KV Bank. The Top-1 KV chunk will be used for Stage 2 retrieval.

\paragraph{Stage 2: page-level retrieval.}

We compute the similarity score between the average query \(\bar{Q} \in \mathbb{R}^{d_k}\) of the current chunk and the average key \(\bar{K}_{m,j} \in \mathbb{R}^{d_k}\) of each KV page from the retrieved chunks in Stage 1:
$\alpha_{m,j} = \frac{\bar{Q}^\top \bar{K}_{m,j}}{\sqrt{d_k}}.$

Based on these scores, we select the top-$K_p$ pages across all candidate chunks to form $\mathcal{P}_{\text{sel}}$. In practice, we set $K_p = 4$ to bound retrieval overhead. 
Since each page contains 8 frames and each frame contains 1560 tokens, 
the retrieved KV cache contains $4 \times 8 \times 1560 = 49{,}920$ tokens.

\paragraph{Active KV construction.}

The final active KV cache is constructed by concatenating the retrieved KV pages:
\begin{equation}
\mathcal{K}_{\text{active}} =
\operatorname{Concat}\bigl\{\mathbf{K}_{m,j}\mid \mathcal{P}_{m,j}\in\mathcal{P}_{\text{sel}}\bigr\},\qquad
\mathcal{V}_{\text{active}} =
\operatorname{Concat}\bigl\{\mathbf{V}_{m,j}\mid \mathcal{P}_{m,j}\in\mathcal{P}_{\text{sel}}\bigr\}.
\end{equation}
These KV pairs are materialized in L1 (GPU HBM)
to guide attention computation of the current chunk.

To manage memory efficiently, KV pages that are not selected by the retrieval stage are automatically migrated to lower memory tiers. When the GPU memory (L1) exceeds its capacity, inactive KV pages are offloaded to CPU memory (L2). If the L2 capacity is further exceeded, the least recently used pages are migrated to the NVMe storage tier (L3). The NVMe tier maintains a full backup of all historical KV pages, ensuring that any page can be restored when needed. As illustrated in Fig. \ref{fig:head_placeholder}, while the total KV cache grows, the GPU memory (L1) occupancy remains bounded by offloading inactive pages to the CPU memory (L2).

\subsection{Hybrid Sparse Attention}
\label{sec:hsa}

Although HKV retrieves relevant pages, they still harbor significant spatial-temporal redundancy. To achieve finer compute allocation, we propose \textit{Hybrid Sparse Attention (HSA)}, a dual-branch sparse attention mechanism that integrates global structure modeling, fine-grained block sparsity, and head-adaptive sparsity control. HSA is designed to reduce attention complexity while preserving expressiveness across disparate attention patterns.

HSA targets video DiTs where a latent volume of shape $(T,H,W)$ is flattened into a sequence of length $L = THW$.
It decomposes attention into a \emph{linear branch} that captures long-range dependencies with linear complexity, a \emph{sparse branch} that performs block-sparse attention over selected regions, and introduces \emph{head-level dynamic sparsity} for per-head compute allocation. 

\paragraph{Linear branch: linear global attention.}
The linear branch applies a sequence-wise projection to reduce the effective attention dimension along the  sequential axis following~\cite{wang2020linformer}. \textbf{Keys} and \textbf{Values} are projected into a lower-dimensional latent space before attention computation, reducing the complexity from $\mathcal{O}(L^2)$ to $\mathcal{O}(Lr)$, which is linear in $L$ when the projection rank $r \ll L$ is fixed. This branch produces a dense but low-rank approximation of global attention, enabling efficient propagation of long-range context without sacrificing coverage.

Concretely, for each head $h$, we introduce two learned projection matrices
$
\mathbf{E}_K^{(h)}\in\mathbb{R}^{L\times r}$ and $\mathbf{E}_V^{(h)}\in\mathbb{R}^{L\times r}$, 
where $r\ll L$ is a fixed projection rank (shared across samples).
We form compressed keys/values
\begin{equation}
\widetilde{\mathbf{K}}^{(h)} 
= \left(\mathbf{E}_K^{(h)}\right)^\top \mathbf{K}^{(h)}, 
\quad \widetilde{\mathbf{K}}^{(h)} \in \mathbb{R}^{r\times d_k}, \quad
\widetilde{\mathbf{V}}^{(h)} 
= \left(\mathbf{E}_V^{(h)}\right)^\top \mathbf{V}^{(h)}, 
\quad \widetilde{\mathbf{V}}^{(h)} \in \mathbb{R}^{r\times d_k}.
\end{equation}

The linear-branch output is then computed via
$\mathbf{O}^{(h)}_{\text{lin}}
=
\text{Softmax}\!\left(\frac{\mathbf{Q}^{(h)}\left(\widetilde{\mathbf{K}}^{(h)}\right)^\top}{\sqrt{d_k}}\right)\widetilde{\mathbf{V}}^{(h)}
\;\in\;\mathbb{R}^{L\times d_k}.$

This reduces attention complexity from $\mathcal{O}(L^2 d_k)$ to $\mathcal{O}(L r d_k)$.

\begin{figure}[t]
\centering

\begin{minipage}[t]{0.32\linewidth}
    \centering
    \includegraphics[width=\linewidth]{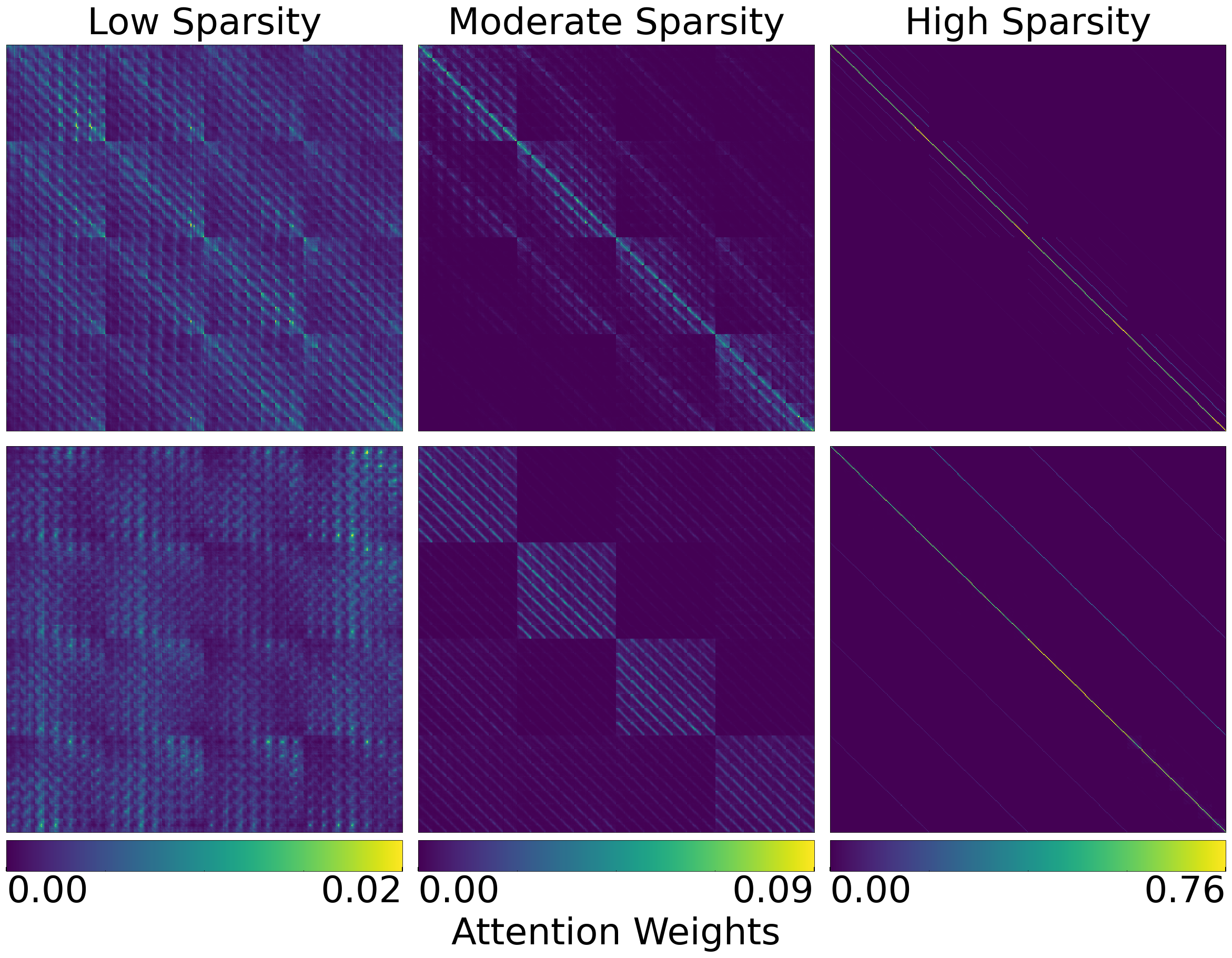}
    \captionof{figure}{Visualization of sparsity patterns across different attention heads.}
    \label{fig:head_map}
\end{minipage}\hfill
\begin{minipage}[t]{0.32\linewidth}
    \centering
    \includegraphics[width=\linewidth]{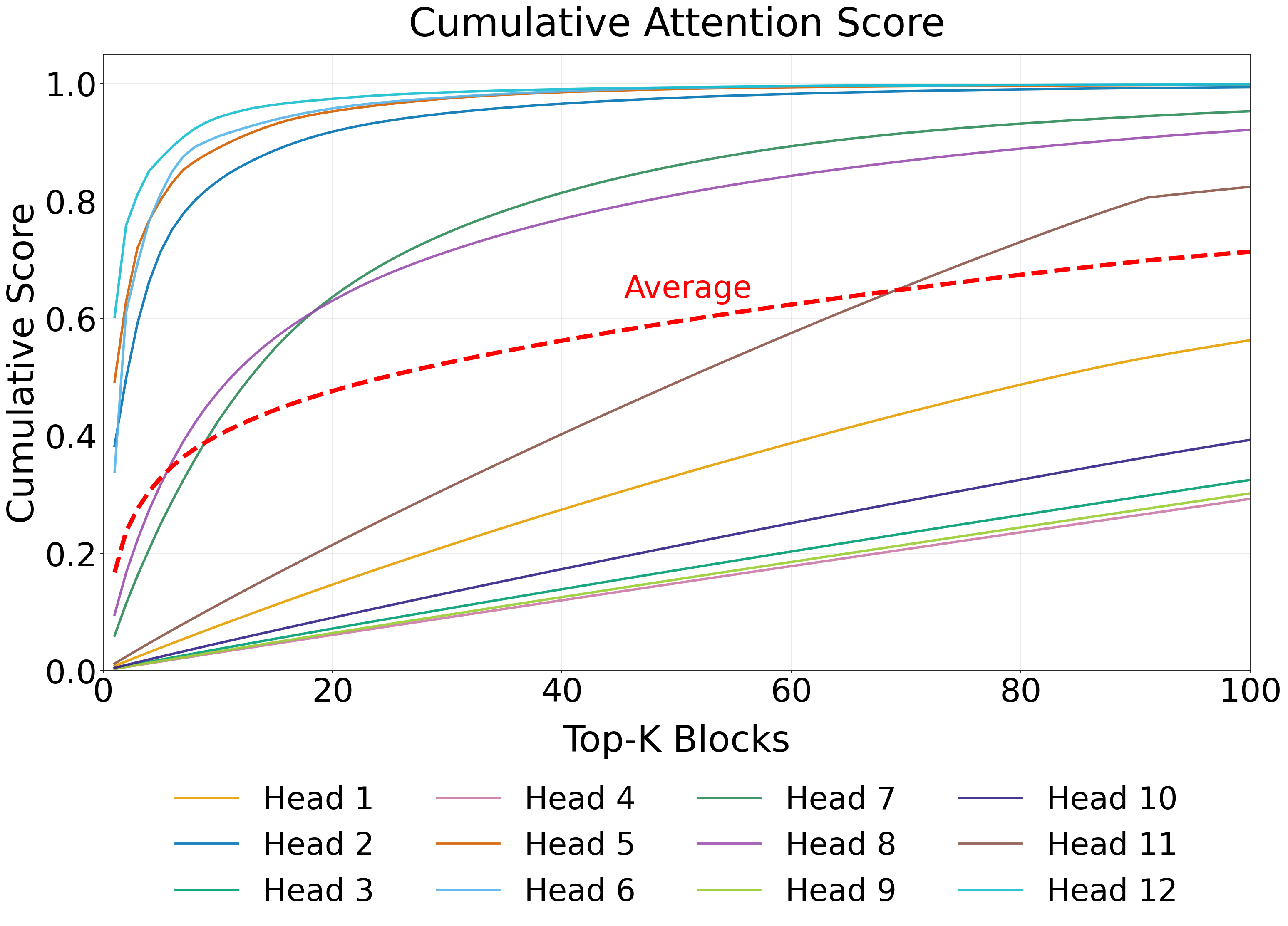}
    \captionof{figure}{Head-wise cumulative attention distribution over Top-$K$ selected blocks.}
    \label{fig:head_curve}
\end{minipage}\hfill
\begin{minipage}[t]{0.32\linewidth}
    \centering
    \includegraphics[width=\linewidth]{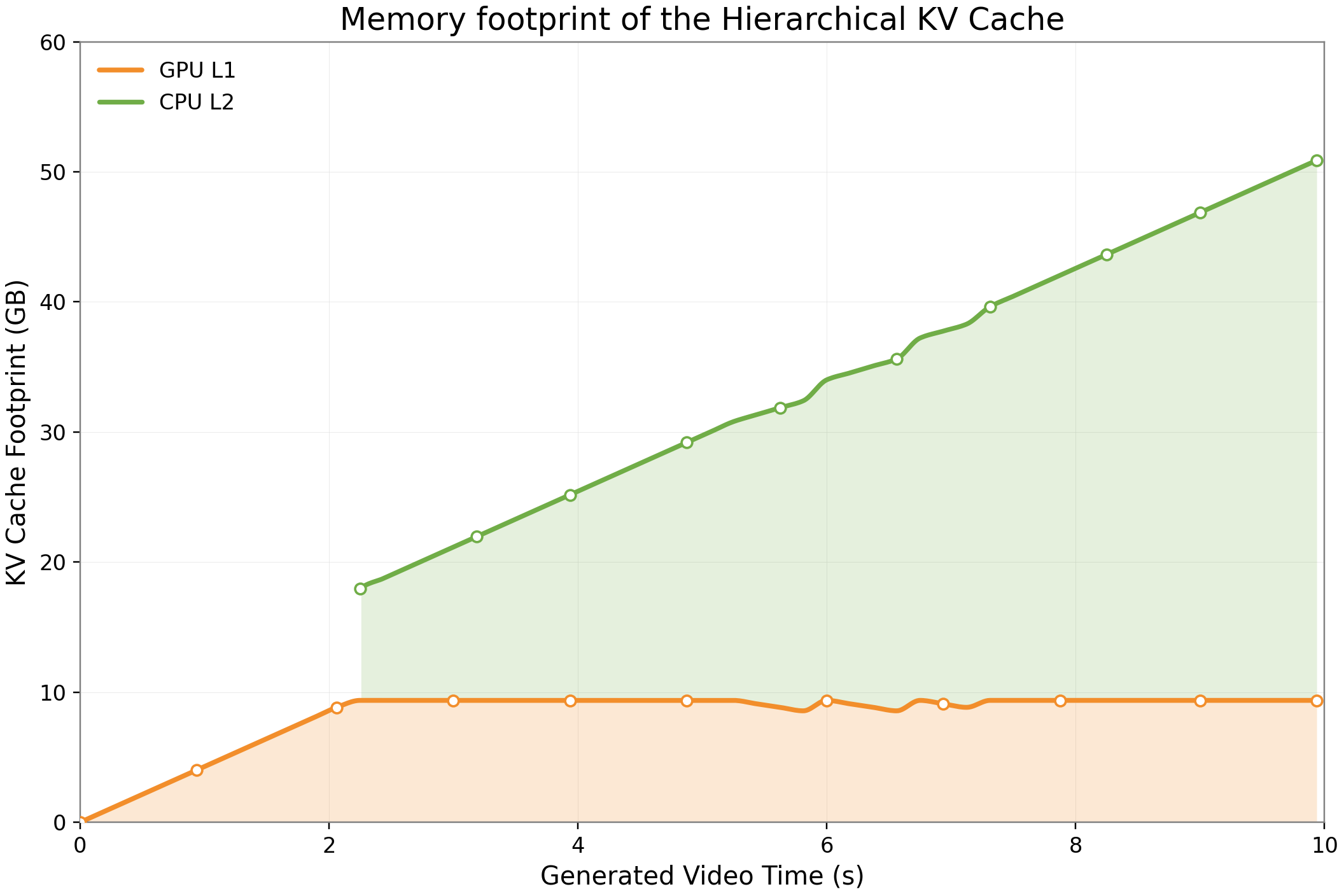}
    \captionof{figure}{Memory footprint analysis of HKV.}
    \label{fig:head_placeholder}
\end{minipage}

\end{figure}

\paragraph{Sparse branch: dynamic block-sparse attention.}
The sparse branch performs attention at the original token resolution but restricts computation to a sparse set of selected blocks. 

We first partition tokens within the retrieved pages into blocks of size $B$ and construct a pooled attention map for coarse block selection. 
Attention is then computed only within the selected blocks using a block-sparse kernel:
\begin{equation}
\mathbf{O}^{(h)}_{\text{sp}}
=
\mathrm{SDPA}_{\text{block}}\!\big(\mathbf{Q}^{(h)},\mathbf{K}^{(h)},\mathbf{V}^{(h)};\,\mathcal{S}^{(h)}\big),
\end{equation}
where $\mathcal{S}^{(h)}$ denotes the set of selected block connections for head $h$.

\emph{Head-level sparsity adaptation}. 
Instead of using a fixed sparsity pattern shared across all heads, we allow each attention head dynamically to determine its own sparsity level based on its attention distribution. 
This is motivated by the observation that different heads capture heterogeneous patterns: some focus on localized interactions, while others encode broader contextual dependencies. As shown in Figure~\ref{fig:head_map}, attention patterns of each head vary from dense global correlations to localized diagonal structures, suggesting a uniform sparsity strategy is suboptimal. 
Moreover, as shown in Figure~\ref{fig:head_curve}, different heads exhibit different accumulation rates, indicating that attention sparsity is head-dependent and motivating our adaptive approach.

Concretely, for each head, we select the minimal set of blocks whose cumulative attention mass exceeds a threshold $\tau_h$, allowing different heads to operate at different sparsity levels and allocate computation adaptively. This head-level adaptation provides finer granularity than layer-level or global sparsity schemes, enabling efficient computation while preserving diverse attention patterns.
The complete algorithmic procedure are provided in Appendix~\ref{sec:hsa_details}.

\emph{Inter- and intra-chunk sparse granularity.}
Following the hardware-aware block-sparse preliminaries in \textit{Appendix~\ref{sec:bsa_detailed}}, 
the tile size $B$ is preferably a multiple of $16$ to align with Hopper MMA / WGMMA computation units. 
This constraint ensures efficient tensor core utilization and avoids degradation in arithmetic intensity.
We distinguish between \textbf{inter-chunk} and \textbf{intra-chunk} sparse granularity. 
For retrieved historical KV pages that serve as auxiliary context, we adopt a coarser partition with 
$B_{\text{inter}}=128$ and $(C_t,C_h,C_w)=(4,8,4)$ to reduce scheduling overhead and improve memory throughput. 
In contrast, tokens within the current chunk are critical for accurate generation, so we employ a finer partition 
$B_{\text{intra}}=64$ with $(C_t,C_h,C_w)=(4,4,4)$ to enable more precise token interactions. 
Both configurations satisfy hardware-friendly divisibility and are executed through the same block-sparse kernel interface, 
balancing computational efficiency and attention precision.

\paragraph{Gated fusion of two branches.}

The outputs of the two branches are combined via a learned gating mechanism applied at the attention-head level.
We combine the two branches using head-specific multiplicative gates applied at the SDPA outputs (without bounding activation).
For each head $h$, we predict gates from the hidden states with two learnable weights $\mathbf{W}^{(h)}_{\text{lin}}$, $\mathbf{W}^{(h)}_{\text{sp}}$, and fuse outputs:
\begin{equation}
\mathbf{G}^{(h)}_{\text{lin}} = \mathbf{X}\mathbf{W}^{(h)}_{\text{lin}},\qquad
\mathbf{G}^{(h)}_{\text{sp}}  = \mathbf{X}\mathbf{W}^{(h)}_{\text{sp}},\qquad
\mathbf{O}^{(h)} = 
\mathbf{O}^{(h)}_{\text{lin}} \odot \mathbf{G}^{(h)}_{\text{lin}}
+
\mathbf{O}^{(h)}_{\text{sp}} \odot \mathbf{G}^{(h)}_{\text{sp}}.
\end{equation}

We also apply a sigmoid activation to the gates~\cite{qiu2025gated}, introducing non-linearity and enabling query-dependent modulation of the two attention branches. This gating mechanism allows the model to dynamically balance the contributions of the linear and sparse branches at both the head and token levels, improving stability and expressiveness. The linear branch ensures global information flow, the sparse branch enables high-resolution modeling where necessary.

\section{Experiments}

\subsection{Benchmarks and Metrics}

We compare our method with state-of-the-art baselines on \textbf{VBench-Long}~\cite{huang2024vbench}. Because the standard VBench-Long protocol is not directly applicable (only 30s), we use the prompts curated by LongLive~\cite{yang2025longlive}, a custom set of 160 interactive 60-second videos, each comprising six successive 10-second prompts.
To showcase our capability in interactive long video generation, we further extend our evaluation to \textbf{InterVBench}~\cite{zhang2025bife,team2025inferix}, which contains 1,000 videos with chunk-level annotations every 2–3 seconds in each video. For metric details, see \textit{Appendix~\ref{sec:metric}}.

\subsection{Implementation Details}
\label{sec:implementation}
To ensure a fair comparison, we follow the training strategy in Self-Forcing~\cite{huang2025self} and LongLive~\cite{yang2025longlive}.
We build WorldAttention upon Wan2.1-T2V-1.3B~\cite{wan2025wan}, which generates 5-second clips at 16 FPS with a resolution of 480p ($832 \times 480$). We first adapt the pretrained model into a 4-step causal-attention model using a self-forcing DMD pipeline~\cite{huang2025self,yin2024one} on the VidProM dataset~\cite{wang2024vidprom}. We then use the student model and a Wan2.1-T2V-14B teacher model to perform \textit{interactive long tuning} on switch prompts with 60s length. The switch prompts are constructed following LongLive~\cite{yang2025longlive}, where Qwen2-72B-Instruct~\cite{bai2023qwen} generates follow-up prompts conditioned on each original VidProM prompt. During training, each iteration extends the model’s own rollout by generating successive 5s clips until reaching a maximum length of 60s. Each training sample contains exactly one prompt switch, with the switch time uniformly sampled between 5s and 55s. The full training process takes approximately 30 hours on 32 NVIDIA H100 GPUs, supported by 192 CPU cores and 960 GB of CPU memory.
We employ AdamW and stepwise decay schedule for all stages of training. The initial learning rate is $1 \times 10^{-4}$, then reduced to $5 \times 10^{-5}$, with the weight decay set to $1 \times 10^{-4}$.

\subsection{Main Results}

\begin{table*}[t]
  \centering
  \caption{\textbf{Comparison of different methods on VBench-Long \cite{huang2024vbench}.} We extended VBench-Long to 60 seconds following LongLive~\cite{yang2025longlive}.}
  \label{tab:vbench}
  \setlength{\tabcolsep}{6pt}
  \renewcommand{\arraystretch}{1.2}
  \resizebox{\linewidth}{!}{
    \begin{tabular}{l|c|c|c|c|c|c}
      \toprule
      Method &
      \makecell{Subject \\ Consistency} $\uparrow$ &
      \makecell{Background \\ Consistency} $\uparrow$ &
      \makecell{Motion \\ Smoothness} $\uparrow$ &
      \makecell{Dynamic \\ Degree} $\uparrow$ &
      \makecell{Aesthetic \\ Quality} $\uparrow$ &
      \makecell{Image \\ Quality} $\uparrow$ \\
      \midrule
      MAGI-1~\cite{teng2025magi} & 0.8320 & 0.8931 & 0.9740 & 0.5537 & 0.5010 & 0.6120\\
      Self Forcing~\cite{huang2025self} & 0.8211 & 0.9050 & 0.9799 & 0.6015 & 0.5130 & 0.6218\\
      PAVDM~\cite{xie2025progressive} & 0.8415 & 0.9273 & 0.9769 & 0.6537 & 0.4970 & 0.6280 \\
      FramePack~\cite{zhang2025packing} & 0.9019 & 0.9450 & 0.9805 & 0.5715 & 0.5044 & 0.6381 \\
      NOVA~\cite{deng2024autoregressive} & 0.7750 & 0.8806 & 0.9894 & 0.1200 & 0.4753 & 0.4497 \\
      CausVid~\cite{yin2025causvid} & 0.8675 & 0.8985 & 0.9847 & 0.5200 & 0.6288 & 0.6747 \\
      Self-Forcing++~\cite{cui2026self} & 0.9165 & 0.9092 & 0.9803 & 0.5865 & 0.5482 & 0.6453 \\
      Deep Forcing~\cite{yi2025deep} & 0.9285 & 0.9136 & 0.9819 & 0.7035 & 0.6041 & 0.6455 \\
      StreamDiffusionV2~\cite{feng2025streamdiffusionv2} & 0.9036 & 0.9051 & 0.9745 & 0.4552 & 0.5547 & 0.5528 \\
      SkyReels-V2-DF-1.3B~\cite{chen2025skyreels} & 0.9391 & 0.9580 & 0.9838 & 0.6529 & 0.5320 & 0.6315 \\
      LCT (MMDiT-3B)~\cite{guo2025long} & 0.9380 & 0.9623 & 0.9816 & 0.6875 & 0.5200 & 0.6345 \\
      MoC~\cite{cai2025mixture} & 0.9398 & 0.9670 & 0.9851 & 0.7500 & 0.5547 & 0.6396 \\
      Infinity-RoPE~\cite{yesiltepe2025infinity} & 0.9352 & 0.9395 & 0.9710 & 0.5395 & 0.6045 & 0.6475 \\
      BIFE~\cite{zhang2025bife} & 0.9410 & 0.9650 & 0.9870 & 0.7720 & 0.5839 & 0.6527 \\
      LongLive~\cite{yang2025longlive} & 0.9403 & 0.9495 & 0.9845 & 0.7321 & 0.5795 & 0.6483 \\
      Rolling Forcing~\cite{liu2025rolling} & 0.9409 & 0.9447 & 0.9865 & 0.3600 & \textbf{0.6350} & 0.7242\\
      \midrule
      \textbf{WorldAttention (Ours)} & \textbf{0.9472} & \textbf{0.9691} & \textbf{0.9915} & \textbf{0.7758} & 0.6344 & \textbf{0.7259}
      \\
      \bottomrule
    \end{tabular}
  }
\vspace{-1em}
\end{table*}

\begin{table}[t]
\centering
\caption{\textbf{Interactive long video evaluation on VBench-Long~\cite{huang2024vbench}.} Quality scores are reported on the full 60s sequence. CLIP scores are reported on 10s video segments with identical semantics ($\uparrow$ higher is better).}
\label{tab:interactive_long}
\setlength{\tabcolsep}{6pt}
\renewcommand{\arraystretch}{1.2}
\resizebox{0.8\linewidth}{!}{
\begin{tabular}{l c cccccc}
\toprule
\multirow{2}{*}{Method} & \multirow{2}{*}{Quality Score $\uparrow$} & \multicolumn{6}{c}{CLIP Score $\uparrow$} \\
\cmidrule(lr){3-8}
 & & 0--10s & 10--20s & 20--30s & 30--40s & 40--50s & 50--60s \\
\midrule
SkyReels-V2 & 80.49 & 20.96 & 22.51 & 25.78 & 18.45 & 19.57 & 19.61 \\
Self-Forcing & 82.46 & 28.46 & 24.89 & 23.53 & 22.96 & 23.07 & 23.19 \\
BIFE & 84.38 & 28.60 & 25.95 & 23.69 & 24.41 & 22.85 & 24.10 \\
LongLive & 83.95 & 28.85 & 25.68 & 24.64 & 24.23 & \textbf{24.32} & 24.32 \\
\midrule
\textbf{WorldAttention (Ours)} & \textbf{85.53} & \textbf{29.04} & \textbf{25.99} & \textbf{25.80} & \textbf{24.72} & 24.25 & \textbf{25.01} \\
\bottomrule
\end{tabular}}
\vspace{-1em}
\end{table}

\paragraph{VBench-Long.} 
As shown in Table~\ref{tab:vbench}, WorldAttention achieves the best performance across nearly all metrics on VBench-Long. 
In particular, it consistently improves subject consistency, background consistency, motion smoothness, and dynamic degree, while also achieving the highest image quality.
For semantic alignment, we split each video at prompt boundaries and compute CLIP-based~\cite{radford2021learning} semantic scores for each segment. As reported in Table~\ref{tab:interactive_long}, WorldAttention demonstrates strong prompt adherence, smooth transitions across prompt changes, and high long-range consistency, while maintaining high generation throughput.
These gains stem from our unified design for long-horizon interactive generation. 
First, HKV preserves full-history context while enabling accurate retrieval, which significantly improves subject and background consistency by preventing identity drift over long rollouts. 
Second, HSA’s dual-branch design keeps global context with its linear branch and captures local details with its sparse branch. This combination improves motion smoothness, preserves important details, and enhances long-range semantic consistency, leading to better visual quality.

\textbf{InterVBench.}~
For results on InterVBench, please refer to Table~\ref{tab:intervbench} in \textit{Appendix~\ref{sec:intervbench}}.

\subsection{Efficiency}

We follow LongLive~\cite{yang2025longlive} to compare efficiency on the single-prompt 30s long video evaluation on VBench-Long~\cite{huang2024vbench}. 
The results in Table~\ref{tab:efficiency_vbench} shows that WorldAttention achieves the highest quality score of \textit{86.55} while also attaining the fastest throughput of \textit{22.0 FPS}, outperforming all previous methods in both generation quality and end-to-end inference efficiency.

We also evaluate the speedup of our kernels in sparse branch attention computation. Our customized kernels \textbf{$HSA_{\text{triton}}$} and \textbf{$HSA_{\text{tk}}$}, which denote the sparse attention kernels implemented in Triton and ThunderKittens, achieve significantly higher speedups compared to the FlashAttention-3 baseline, as shown in Fig.~\ref{fig:kernel_speedup}. When accumulating the runtime across the video generation, \textbf{$HSA_{\text{tk}}$} delivers a peak speedup of \textbf{14.02x}, validating our system-oriented co-design of sparse attention.

\begin{table}[t]
\centering
\small

\begin{minipage}{0.55\linewidth}
\vspace{-1cm}
\centering
\caption{Single-prompt 30s long video evaluation on VBench-Long~\cite{huang2024vbench}.}
\label{tab:efficiency_vbench}
\resizebox{\linewidth}{!}{
\begin{tabular}{lcc}
\toprule
Model & Quality Score $\uparrow$ & Throughput (FPS) $\uparrow$ \\
\midrule
SkyReels-V2 & 80.77 & 0.49 \\
FramePack & 83.61 & 0.92 \\
Self-Forcing & 83.82 & 17.0 \\
BIFE & 85.18 & 18.0 \\
LongLive (re-cache 12 frames) & 85.44 & 20.7 \\
LongLive (re-cache 32 frames) & 85.82 & 6.70 \\
\midrule
\textbf{WorldAttention (Ours)} & \textbf{86.55} & \textbf{22.0} \\
\bottomrule
\end{tabular}
}
\end{minipage}
\hfill
\begin{minipage}{0.36\linewidth}
    \vspace{-1.5em} %
    \centering
    \includegraphics[width=\linewidth]{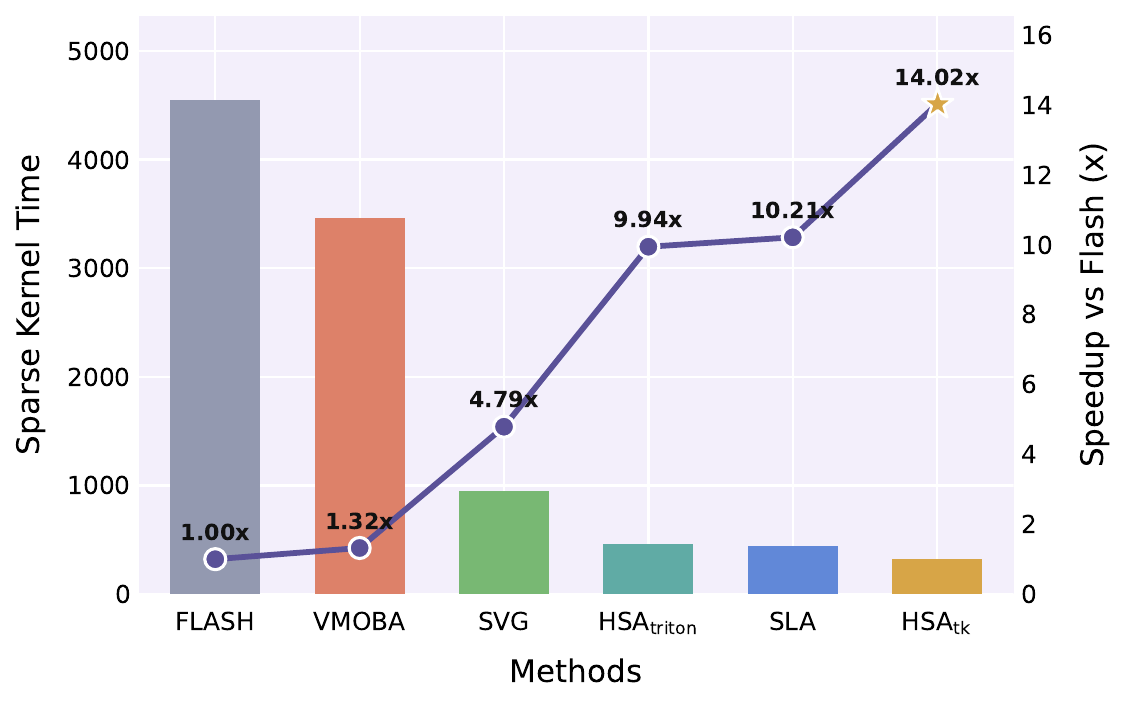}
    \vspace{-1em}
    \captionof{figure}{Attention kernel speedup.}
    \label{fig:kernel_speedup}
\end{minipage}
\vspace{-1.5em}
\end{table}

\subsection{Ablation Studies}
\label{sec:ablations}

\begin{table*}[t]
  \centering
  \caption{\textbf{Comparison of different modification on HSA.} We report VBench-Long~\cite{huang2024vbench} metrics following~\cite{guo2025long,cai2025mixture}.}
  \label{tab:hsa}
  \setlength{\tabcolsep}{6pt}
  \renewcommand{\arraystretch}{1.2}
  \resizebox{\linewidth}{!}{
    \begin{tabular}{l|c|c|c|c|c|c}
      \toprule
      Method &
      \makecell{Subject \\ Consistency} $\uparrow$ &
      \makecell{Background \\ Consistency} $\uparrow$ &
      \makecell{Motion \\ Smoothness} $\uparrow$ &
      \makecell{Dynamic \\ Degree} $\uparrow$ &
      \makecell{Aesthetic \\ Quality} $\uparrow$ &
      \makecell{Image \\ Quality} $\uparrow$ \\
      \midrule
      Linear Only & 0.9021 & 0.9365 & 0.9747 & 0.5825 & 0.5632 & 0.6490
      \\
      \midrule
      \multicolumn{7}{l}{Sparse Only:} \\
      \textbf{$\tau_{\max} = 0.75, \tau_{\min} = 0.35$} & 0.8736 & 0.8845 & 0.9735 & 0.6053 & 0.6342 & 0.7042
      \\
      \textbf{$\tau_{\max} = 1.00, \tau_{\min} = 0.10$} & 0.9254 & 0.9424 & 0.9880 & 0.6621 & 0.6293 & 0.7035
      \\
      \textbf{$\tau_{\max} = 1.00, \tau_{\min} = 0.35$} & 0.9365 & 0.9486 & 0.9894 & 0.6945 & 0.6046 & 0.6995
      \\
      \midrule
      \multicolumn{7}{l}{Sparse and Linear:} \\
      \textbf{$\tau_{\max} = 0.75, \tau_{\min} = 0.35$} & 0.9300 & 0.9270 & 0.9863 & 0.7520 & 0.6201 & 0.7015
      \\
      \textbf{$\tau_{\max} = 1.00, \tau_{\min} = 0.10$} & 0.9415 & 0.9593 & 0.9914 & 0.7507 & 0.6287 & 0.7148
      \\
      \rowcolor{Periwinkle!30}\textbf{$\tau_{\max} = 1.00, \tau_{\min} = 0.35$} & \textbf{0.9472} & \textbf{0.9691} & \textbf{0.9915} & \textbf{0.7758} & \textbf{0.6344} & \textbf{0.7259}
      \\
      \bottomrule
    \end{tabular}
  }
\end{table*}

\begin{figure}[t!]
    \centering
    \begin{subfigure}{0.32\textwidth}
        \centering
        \includegraphics[width=\textwidth]{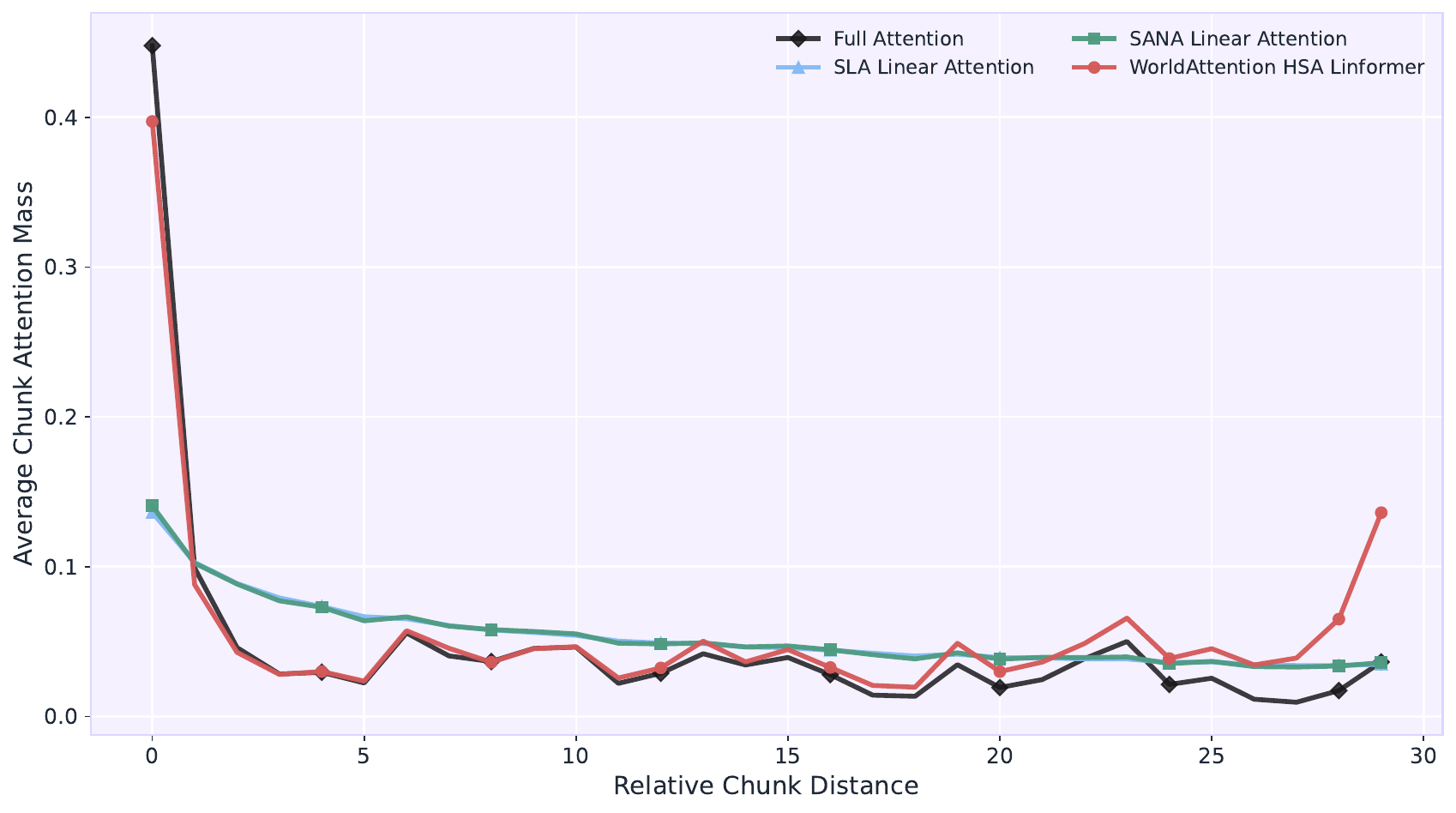}
        \caption{Global attention mass.}
        \label{fig:attn_mass}
    \end{subfigure}
    \hfill
    \begin{subfigure}{0.32\textwidth}
        \centering
        \includegraphics[width=\textwidth]{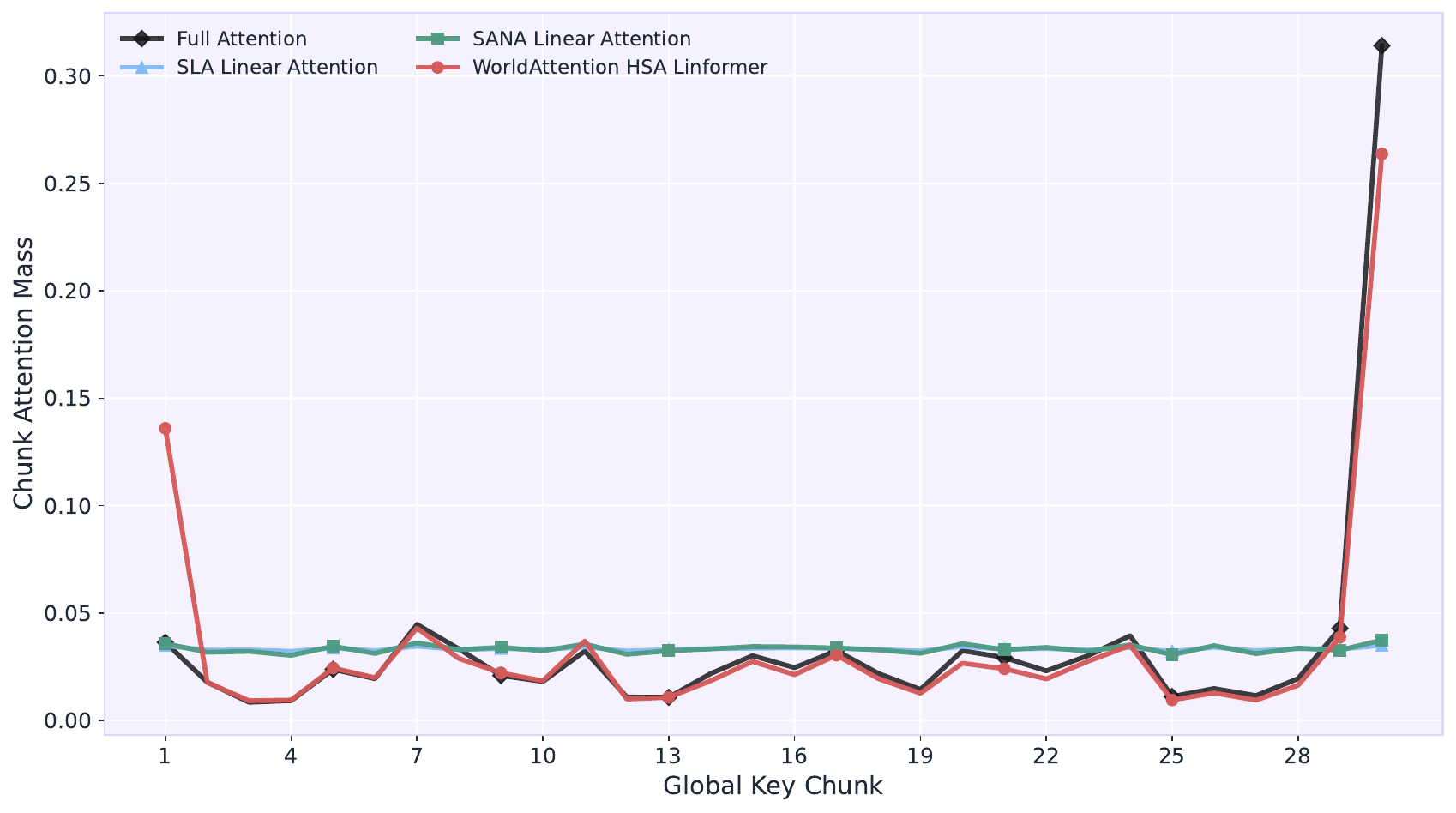}
        \caption{Distribution of last chunk.}
        \label{fig:attn_chunk}
    \end{subfigure}
    \hfill
    \begin{subfigure}{0.32\textwidth}
        \centering
        \includegraphics[width=\textwidth]{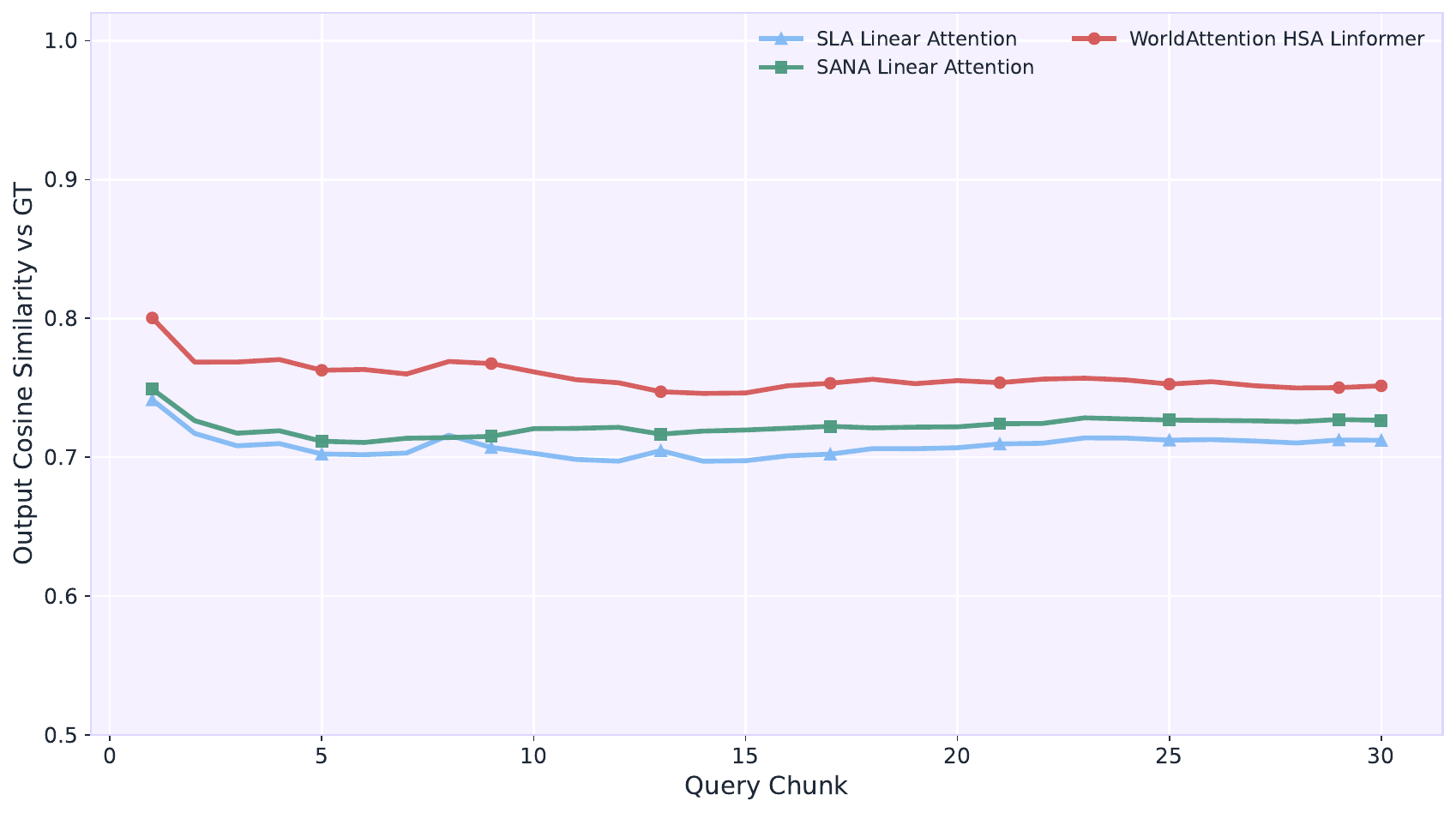}
        \caption{Output cosine similarity.}
        \label{fig:cosine_sim}
    \end{subfigure}
    
    \caption{Comparison of {\color{red!80!black}our linear branch} with \textcolor{NavyBlue}{SLA}~\cite{zhang2025sla}, \textcolor{PineGreen}{SANA-Video}~\cite{chen2025sana}, and full attention.}
    \label{fig:attn_analysis}
    \vspace{-2em}
\end{figure}

\paragraph{Hybrid sparse attention.}

Table~\ref{tab:hsa} compares different design choices for HSA on VBench-Long. 
Using only the linear branch yields strong subject and background consistency but produces weaker motion dynamics and visual quality. 
Sparse attention alone improves dynamic degree and aesthetic quality, while its performance depends on the sparsity thresholds $(\tau_{\max}, \tau_{\min})$. 
Combining sparse and linear branches consistently improves performance across most metrics, indicating that the two branches provide complementary modeling capacity. 
In particular, the configuration $(\tau_{\max}=1.00, \tau_{\min}=0.35)$ achieves the best overall results, reaching the highest scores in subject consistency, background consistency, motion smoothness, aesthetic quality, and image quality while maintaining strong motion dynamics.
Moreover, we compare our \textbf{linear branch} with SLA~\cite{zhang2025sla}, SANA-Video~\cite{chen2025sana}, and full attention within the WorldAttention architecture. As shown in Fig.~\ref{fig:attn_mass}, our linear branch accurately recovers the long-range attention mass of full attention and, as shown in Fig.~\ref{fig:attn_chunk}, maintains consistent distribution patterns across specific chunks. Fig.~\ref{fig:cosine_sim} shows that our linear branch achieves superior output embedding similarity to the full-attention baseline.
For more analysis on linear attention, see \textit{Appendix~\ref{sec:linear}}.

\paragraph{Hierarchical KV cache.}

Table~\ref{tab:hkv} presents an ablation study of the proposed HKV on interactive long-video generation. 
Sliding-window caching shows clear degradation in CLIP scores over time due to limited temporal context. 
Re-caching more frames improves semantic consistency but significantly reduces throughput, highlighting the efficiency--quality trade-off. 
In contrast, HKV maintains high throughput while improving both the overall quality score and segment-wise CLIP scores. 
Among different configurations, HKV with $(P_{size}=8, K_p=4)$ achieves the best overall performance, delivering the highest quality score and the strongest long-horizon semantic consistency while preserving competitive inference speed.

\begin{table}[t]
\centering
\caption{\textbf{Ablation of HKV: interactive long video evaluation on VBench-Long~\cite{huang2024vbench}.} Quality scores are reported on the full 60s sequence. CLIP scores are reported on 10s video segments with identical semantics ($\uparrow$ higher is better).}
\label{tab:hkv}
\setlength{\tabcolsep}{6pt}
\renewcommand{\arraystretch}{1.2}
\resizebox{\linewidth}{!}{
\begin{tabular}{l c c cccccc}
\toprule
\multirow{2}{*}{Method} & \multirow{2}{*}{Throughput (FPS) $\uparrow$} & \multirow{2}{*}{Quality Score $\uparrow$} & \multicolumn{6}{c}{CLIP Score $\uparrow$} \\
\cmidrule(lr){4-9}
 & & & 0--10s & 10--20s & 20--30s & 30--40s & 40--50s & 50--60s \\
\midrule
Sliding Window & 22.1 & 82.52 & 25.46 & 22.74 & 21.49 & 19.94 & 18.45 & 17.59 \\
Sliding Window (re-cache 12 frames) & 20.8 & 84.05 & 28.40 & 24.89 & 24.07 & 23.19 & 22.95 & 23.64 \\
Sliding Window (re-cache 32 frames) & 7.10 & 84.82 & 28.52 & 24.45 & 25.26 & 24.05 & 23.71 & 24.47 \\
\midrule
HKV ($P_{size} = 2, K_p = 16$) & 21.9 & 85.38 & 29.01 & 25.51 & 25.37 & 24.69 & 24.19 & 24.93 \\
HKV ($P_{size} = 4, K_p = 8$) & 21.9 & 84.93 & 24.81 & \textbf{26.09} & 25.63 & 24.64 & 24.07 & 24.10 \\
HKV ($P_{size} = 8, K_p = 2$) & \textbf{24.0} & 83.05 & 27.48 & 25.08 & 25.12 & 24.10 & 23.99 & 24.28 \\
\rowcolor{Periwinkle!30}HKV ($P_{size} = 8, K_p = 4$) & 22.0 & \textbf{85.53} & \textbf{29.04} & 25.99 & \textbf{25.80} & \textbf{24.72} & \textbf{24.25} & \textbf{25.01} \\
\bottomrule
\end{tabular}}
\vspace{-1em}
\end{table}

\begin{table*}[t!]
\centering
\caption{Comparison of hierarchical KV retrieval with alternative page-selection strategies under the same active KV budget of four pages (49,920 tokens). Quality is evaluated on the full 60-second VBench-Long sequence; CLIP scores are evaluated on aligned 10-second segments.}
\label{tab:hkv_retrieval_ablation}
\renewcommand{\arraystretch}{1.15}
\scriptsize
\resizebox{\linewidth}{!}{%
\begin{tabular}{@{}lcccccccc@{}}
\toprule
Method
& Throughput (FPS) $\uparrow$
& Quality $\uparrow$
& \makecell{CLIPScore\\0--10s}
& \makecell{CLIPScore\\10--20s}
& \makecell{CLIPScore\\20--30s}
& \makecell{CLIPScore\\30--40s}
& \makecell{CLIPScore\\40--50s}
& \makecell{CLIPScore\\50--60s} \\
\midrule
Page-only retrieval over all history
& 19.8 & 85.21 & 28.97 & 25.84 & 25.60 & 24.58 & 24.06 & 24.78 \\
Prompt-only chunk retrieval
& 21.9 & 84.79 & 28.76 & 25.48 & 25.12 & 24.10 & 23.74 & 24.05 \\
Chunk-level retrieval
& \textbf{22.3} & 84.48 & 28.41 & 25.21 & 24.86 & 23.88 & 23.45 & 23.70 \\
Recent-page retrieval
& 22.1 & 83.74 & 27.93 & 24.72 & 23.92 & 22.91 & 22.05 & 21.48 \\
Random-page retrieval
& 21.4 & 82.86 & 26.85 & 23.71 & 22.64 & 21.46 & 20.63 & 20.12 \\
Mismatched-page retrieval
& 22.0 & 80.96 & 25.92 & 22.46 & 20.91 & 19.27 & 18.14 & 17.36 \\
\midrule
\rowcolor{Periwinkle!30}\textbf{HKV (Ours)}
& 22.0 & \textbf{85.53} & \textbf{29.04} & \textbf{25.99}
& \textbf{25.80} & \textbf{24.72} & \textbf{24.25} & \textbf{25.01} \\
\bottomrule
\end{tabular}%
}
\end{table*}

\begin{table*}[t]
\centering
\caption{Ablation of the number of prompt-level candidate chunks under the same active KV budget of four pages.}
\label{tab:topk_chunk_ablation}
\renewcommand{\arraystretch}{1.15}
\scriptsize
\resizebox{\linewidth}{!}{%
\begin{tabular}{@{}lcccccccc@{}}
\toprule
Method
& Throughput (FPS) $\uparrow$
& Quality $\uparrow$
& \makecell{CLIPScore\\0--10s}
& \makecell{CLIPScore\\10--20s}
& \makecell{CLIPScore\\20--30s}
& \makecell{CLIPScore\\30--40s}
& \makecell{CLIPScore\\40--50s}
& \makecell{CLIPScore\\50--60s} \\
\midrule
Top-4 chunks, 1 page per chunk
& 20.4 & 84.91 & 28.72 & 25.42 & 25.21 & 24.18 & 23.77 & 24.13 \\
Top-2 chunks, 2 pages per chunk
& 21.2 & 85.31 & 28.93 & 25.78 & 25.57 & 24.55 & 24.09 & 24.72 \\
\rowcolor{Periwinkle!30}\textbf{Top-1 chunk, 4 pages (Ours)}
& \textbf{22.0} & \textbf{85.53} & \textbf{29.04} & \textbf{25.99}
& \textbf{25.80} & \textbf{24.72} & \textbf{24.25} & \textbf{25.01} \\
\bottomrule
\end{tabular}%
}
\end{table*}

\paragraph{Ablation of Retrieval Design.}

We use hierarchical retrieval because the two stages solve different
problems: (1) page-only retrieval must compare against every page in the
growing history, causing retrieval cost to increase with video length; and
(2) prompt-only chunk retrieval is cheap but too coarse, as it loads an
entire chunk even when only a few pages are relevant. HKV first uses prompt
similarity to efficiently narrow the search space, and then applies
query--key similarity to select only the relevant pages. This design provides
both scalable retrieval and fine-grained context selection.
To further clarify this analysis, we compare HKV with simpler alternatives on interactive long-video
generation with VBench-Long. Quality scores are evaluated on the full
60-second sequence, whereas CLIP scores are reported on 10-second video
segments with identical semantics.
As shown in Table~\ref{tab:hkv_retrieval_ablation}, under the same four-page
active KV budget, HKV achieves a quality score of 85.53, compared with 83.74
for recent-page retrieval, 82.86 for random-page retrieval, and 80.96 for
mismatched-page retrieval. The advantage is more pronounced in the
50--60s segment, where HKV achieves a CLIP score of 25.01, compared with
21.48, 20.12, and 17.36, respectively. Since these methods differ only in
their page-selection strategy, the results indicate that the query--key score
retrieves meaningful visual memory rather than arbitrary, merely recent, or
deliberately mismatched pages.

\paragraph{Ablation of Prompt-Level Candidate Chunks.}
Top-$k$ chunk retrieval could be more robust for genuinely multi-event
prompts. However, under our current setting, retrieving more chunks dilutes the fixed page budget and introduces more irrelevant context. Since each evaluation sample mainly targets one semantic chunk, Top-1 provides the best balance between retrieval precision and efficiency. To study this design, we compare different numbers of prompt-level candidate chunks under the same active KV budget of four pages (49,920 tokens). Specifically, Top-1 retrieves four pages from one chunk, Top-2 retrieves two pages from each chunk, and Top-4 retrieves one page from each chunk. The results are reported in Table~\ref{tab:topk_chunk_ablation}. We conduct the experiment as an interactive long-video evaluation on VBench-Long. Quality scores are reported on the full 60-second sequence, whereas CLIP scores are reported on 10-second video segments with identical semantics. Top-4 retrieval covers more semantic chunks, but allocating only one page to each chunk provides insufficient fine-grained visual information and increases retrieval overhead. Top-2 retrieval improves candidate diversity, but splitting the page budget across two chunks can introduce less relevant context and reduce within-chunk visual coverage. In contrast, Top-1 retrieval (ours) concentrates the fixed KV budget on the most relevant semantic chunk, enabling
more complete fine-grained retrieval while maintaining the highest throughput.

\begin{table*}[t]
\centering
\caption{Robustness of hierarchical retrieval to semantically similar historical chunks. All settings use the same active KV budget of four pages (49,920 tokens).}
\label{tab:similar_chunk_ablation}
\renewcommand{\arraystretch}{1.15}
\scriptsize
\resizebox{\linewidth}{!}{%
\begin{tabular}{@{}lcccccccc@{}}
\toprule
Setting
& Throughput (FPS) $\uparrow$
& Quality $\uparrow$
& \makecell{CLIPScore\\0--10s}
& \makecell{CLIPScore\\10--20s}
& \makecell{CLIPScore\\20--30s}
& \makecell{CLIPScore\\30--40s}
& \makecell{CLIPScore\\40--50s}
& \makecell{CLIPScore\\50--60s} \\
\midrule
\rowcolor{Periwinkle!30}No manually constructed similar chunks
& \textbf{22.0} & \textbf{85.53} & \textbf{29.04} & \textbf{25.99}
& \textbf{25.80} & \textbf{24.72} & \textbf{24.25} & \textbf{25.01} \\
2 manually constructed similar chunks
& 21.9 & 85.36 & 28.96 & 25.91 & 25.69 & 24.60 & 24.13 & 24.88 \\
3 manually constructed similar chunks
& 21.9 & 85.18 & 28.87 & 25.80 & 25.56 & 24.47 & 24.01 & 24.74 \\
\bottomrule
\end{tabular}%
}
\end{table*}

\paragraph{Robustness to Semantically Similar Historical Chunks.}
Semantically similar historical prompts present a challenging retrieval setting.
However, similar prompt embeddings do not necessarily cause retrieval failure,
as the two stages operate at complementary levels: prompt-level retrieval
identifies the most relevant semantic chunk, while page-level query--key
similarity selects fine-grained visual evidence within that chunk. To evaluate
this robustness, we construct a challenging subset of 200 prompts in which
multiple historical chunks have high prompt similarity to the current prompt.
Under the same active KV budget of four pages, Table~\ref{tab:similar_chunk_ablation}
shows only a small performance drop as the number of semantically similar
chunks increases. In particular, quality decreases from 85.53 with no
manually constructed similar chunks to 85.18 with three such chunks, while
the CLIP score for the 50--60s segment remains 24.74. These results indicate
that the two-stage retrieval mechanism remains robust under semantic
ambiguity.

\begin{table}[t]
  \centering

  \begin{minipage}[t]{0.48\linewidth}
\centering
\small
\caption{End-to-end ablation study of WorldAttention.}
\label{tab:e2e_ablation}
\begin{tabular}{lc}
\toprule
System Configuration & E2E Speedup \\
\midrule
Baseline (FA-3 + Naive KV) & 1.00$\times$ \\
+ Hierarchical KV Cache (HKV) & 1.15$\times$ \\
+ Hybrid Sparse Attention (HSA) & 1.91$\times$ \\
+ Kernel Customization & 2.21$\times$ \\
\bottomrule
\end{tabular}
  \end{minipage}\hfill
  \begin{minipage}[t]{0.48\linewidth}
    \centering
    \small
    \captionof{table}{Execution time breakdown of HSA.}
    \label{tab:hsa_breakdown}
    \resizebox{0.8\linewidth}{!}{
    \begin{tabular}{lcc}
    \toprule
    Stage & Time ($\mu$s) & Ratio \\
    \midrule
    Linear branch & 251.934 & 39.78\% \\
    Dynamic selection & 83.359 & 13.16\% \\
    Index mapping (pre-kernel) & 79.135 & 12.50\% \\
    KV contiguous copy & 196.254 & 31.00\% \\
    Sparse block attention kernel & 11.776 & 1.86\% \\
    Fusion & 10.784 & 1.70\% \\
    \midrule
    Total & 633.242 & 100\% \\
    \bottomrule
    \end{tabular}}
  \end{minipage}
\vspace{-1em}
\end{table}

\paragraph{Kernel speedup.}
Despite the substantial kernel-level acceleration, the sparse attention kernel itself contributes only a small fraction of the overall runtime. Table~\ref{tab:hsa_breakdown} shows the execution breakdown of HSA during decoding. The sparse block attention kernel accounts for only $1.86\%$ of total latency, while preprocessing steps such as KV cache reorganization and contiguous memory copies dominate the execution time. This observation suggests that kernel optimization alone cannot fully address the efficiency bottleneck of long-context video generation. Instead, system-level co-design of sparse attention and KV cache management is necessary to achieve substantial end-to-end speedup.

\paragraph{End-to-End Efficiency.}
We also conducted an end-to-end ablation study on minute-long video generation. Even though our customized kernels exclusively accelerate the denoising stage, they still translate into highly competitive overall speedups. As Table~\ref{tab:e2e_ablation} shows, integrating HKV and HSA into the baseline progressively yields $1.15\times$ and $1.91\times$ speedups by reducing I/O and theoretical FLOPs. Finally, our system-oriented kernel 
customization achieve a $2.21\times$ overall end-to-end speedup, confirming that hardware-algorithm co-design is essential for scalable long-video generation.

\begin{wraptable}{r}{0.6\textwidth}
\centering
\caption{End-to-end speedup across GPU platforms, model scales, resolutions, and generation horizons.}
\label{tab:efficiency_generalization}
\renewcommand{\arraystretch}{1.15}
\scriptsize
\resizebox{\linewidth}{!}{%
\begin{tabular}{@{}lcccc@{}}
\toprule
\makecell[l]{System\\configuration}
& \makecell{H100, 1.3B\\480p, 60s}
& \makecell{B200, 1.3B\\480p, 60s}
& \makecell{B200, 1.3B\\480p, 90s}
& \makecell{B200, 14B\\720p, 60s} \\
\midrule
Baseline (FA-3)
& 1.00$\times$ & 1.00$\times$ & 1.00$\times$ & 1.00$\times$ \\
+ HKV
& 1.15$\times$ & 1.12$\times$ & 1.21$\times$ & 1.16$\times$ \\
+ HSA
& 1.91$\times$ & 1.79$\times$ & 2.04$\times$ & 2.13$\times$ \\
+ Kernel Customization
& \textbf{2.21$\times$} & \textbf{2.06$\times$}
& \textbf{2.35$\times$} & \textbf{2.46$\times$} \\
\bottomrule
\end{tabular}%
}
\end{wraptable}

\paragraph{Efficiency Across Hardware, Model Scales, and Generation Settings.}
We further evaluate the end-to-end efficiency gains of WorldAttention on the latest NVIDIA B200 GPU, which also supports our kernels, across different model scales, resolutions, and a longer 90-second generation horizon. The results are reported in Table~\ref{tab:efficiency_generalization}. These results show that WorldAttention maintains consistent end-to-end efficiency gains on B200 across model scales, resolutions, and generation horizons. The benefits become more pronounced for longer sequences and larger models, where KV management and attention computation account for a greater fraction of inference cost.

\section{Conclusion}

In this paper, we introduce WorldAttention, a system-oriented framework for efficient text-conditioned interactive video world models. We present Hybrid Sparse Attention (HSA), which combines linear global attention with head-adaptive block sparsity to reduce quadratic attention complexity while preserving expressive token interactions. 
Furthermore, we develop a Hierarchical KV Cache (HKV) to alleviate the memory pressure of extended sequences, utilizing a multi-tier paging strategy that dynamically balances context  with restricted GPU resources. Through the %
integration of attention structure, memory management, and hardware-aligned kernels, WorldAttention transforms sparse attention and KV retrieval into an efficient execution pipeline for long-horizon video generation. Extensive experiments on VBench-Long and InterVBench demonstrate that our approach achieves superior temporal consistency, prompt alignment, and generation quality while significantly improving decoding efficiency.

\bibliographystyle{plain}
\bibliography{paper}

\clearpage
\appendix

\section{Preliminaries}
\label{sec:preliminaries}

\subsection{Block Sparse Attention}
\label{sec:bsa}

\subsubsection{3D Full Attention}
Modern video Diffusion Transformers (DiTs) typically rely on \emph{3D full attention} to model dependencies across the entire video volume. Given a video latent tensor of shape $(T,H,W)$, the latent is flattened into a 1D token sequence of length $L = THW$. Each token at spatial-temporal location $(t,h,w)$ is mapped to a unique index $n$ in the flattened sequence via:
\[
n = tHW + hW + w.
\]
Full attention is then applied over this sequence, allowing every token to attend to all others.

Formally, for a single attention head, let $\mathbf{Q}, \mathbf{K}, \mathbf{V} \in \mathbb{R}^{L \times d}$ denote the query, key, and value matrices, and let $\mathbf{M} \in \{-\infty,0\}^{L \times L}$ be an attention mask that specifies permissible token interactions. The attention output is computed as
\begin{equation}
\mathbf{S} = \frac{\mathbf{Q}\mathbf{K}^\top}{\sqrt{d_k}}, \quad
\mathbf{A} = \mathrm{Softmax}(\mathbf{S} + \mathbf{M}), \quad
\mathbf{O} = \mathbf{A}\mathbf{V}.
\label{eq:attention}
\end{equation}
In \emph{full attention}, all entries of $\mathbf{M}$ are zero, resulting in quadratic complexity with respect to the sequence length.

\subsubsection{Sparse Attention}
Empirical studies~\cite{he2025zipvl} have shown that attention score matrix $\mathbf{A}$ is inherently sparse, with most entries close to zero. This observation motivates selectively preserving only the high-magnitude regions of $\mathbf{A}$, referred to as \emph{critical tokens}, while discarding the rest. A key design challenge is determining how much computation should be devoted to identifying these tokens.

Computing full attention scores yields the most accurate selection but largely eliminates computational savings, as sparsity only benefits the $\mathbf{A}\mathbf{V}$ stage. In contrast, fixed sparsity patterns introduce no additional prediction overhead, but often fail to capture semantically important long-range interactions. A promising middle ground is to employ a lightweight, trainable, coarse-grained attention mechanism that estimates the locations of critical tokens without explicitly computing the full attention matrix. The central challenge lies in achieving an effective balance between selection accuracy and computational overhead in practical DiT architectures.

\textit{Sparse attention}~\cite{zhang2025flashmo} reduces computation by setting a subset of entries in $\mathbf{M}$ to $-\infty$, thereby skipping the corresponding interactions in both $\mathbf{Q}\mathbf{K}^\top$ and $\mathbf{A}\mathbf{V}$. While this strategy reduces theoretical FLOPs, unstructured sparsity is poorly matched to modern GPU architectures, which are optimized for dense matrix operations and thus fail to realize practical speedups.

\subsubsection{Block Sparse Attention}
\label{sec:bsa_detailed}

\textit{Block-sparse attention}~\cite{xi2025sparse,yang2025sparse,xu2025xattention} addresses this limitation by enforcing structured sparsity aligned with hardware execution. A \textit{block} in GPU refer to a submatrix that a GPU threadblock loads into SRAM when performing matrix multiplication. Specifically, the attention mask $\mathbf{M}$ is partitioned into blocks $(B_q,B_k)$, where all entries within a block share the same mask value. Technically, $(B_q, B_k)$ need not be square and may take different values for $B_q$ and $B_k$. For simplicity, we assume square blocks with $B = B_q = B_k$. This design enables each block to be processed as a dense block or skipped entirely, allowing efficient execution within GPU streaming multiprocessors (SM).

The \textit{block size} $B$ plays a critical role in balancing expressiveness and efficiency. Smaller blocks permit finer-grained attention patterns but reduce hardware utilization, whereas larger blocks improve throughput at the cost of coarser attention resolution. In practice, modest sacrifices in raw speed are often acceptable when they lead to meaningful gains in generation quality. Moreover, the FlashAttention-3~\cite{shah2024flashattention} kernel can support relatively small block sizes. However, it is optimized for Hopper GPUs. On Hopper, MMA and WGMMA instructions use $16 \times 16$ (or their compositions) as the fundamental computation units. As a result, the block size $B$ is preferably divisible by $16$.

Given a video latent tensor of shape $(T,H,W)$, block-sparse attention partitions the volume into a set of contiguous 3D cubes, each of size $(C_t, C_h, C_w)$. The algorithm and its kernel implementation are co-designed such that each cube is mapped to a single execution block on a GPU SM. The resulting block size is
\[
B = C_t \times C_h \times C_w,
\]
corresponding to the number of tokens within one cube.

We assume that $(T,H,W)$ is an integer multiple of $(C_t,C_h,C_w)$, and define the number of cubes along each dimension as
\[
(N_t, N_h, N_w) = \left(\frac{T}{C_t}, \frac{H}{C_h}, \frac{W}{C_w}\right).
\]
When flattening the 3D latent into a one-dimensional sequence, a token located at $(t,h,w)$ is assigned an index $n$ according to

\[
\begin{aligned}
n
&=
\Big(
\Big\lfloor \tfrac{t}{C_t} \Big\rfloor N_h N_w
+
\Big\lfloor \tfrac{h}{C_h} \Big\rfloor N_w
+
\Big\lfloor \tfrac{w}{C_w} \Big\rfloor
\Big) B \\
&\quad
+
(t \bmod C_t) C_h C_w
+
(h \bmod C_h) C_w
+
(w \bmod C_w).
\end{aligned}
\]

This indexing scheme guarantees that every contiguous group of $B$ tokens in the flattened sequence is assigned to the same GPU block and corresponds to a spatially and temporally contiguous region in the original video volume, namely a $(C_t,C_h,C_w)$ cube. As a result, block-level sparsity in attention directly translates to skipping or executing entire GPU blocks, enabling efficient hardware-aligned sparse computation.

\subsubsection{Global and Local Context in Sparse Attention}
Restricting attention connectivity inevitably limits the receptive field, which can impair the modeling of global context. One strategy to mitigate this issue is to augment sparse attention with a lightweight global module that captures coarse, long-range information~\cite{zhang2025sla,lee2023sea,zhang2025vsa}. Alternatively, incorporating local inductive biases, motivated by the success of locality priors in convolutional networks, can enhance fine-grained feature learning~\cite{zhang2025fast}.

\subsection{Semantic Sparse KV Cache}
\label{sec:semantic_sparse}

Long-horizon video generation requires conditioning on historical context through KV caches. 
A naive strategy stores all previous key–value states, causing rapidly growing memory usage 
and redundant attention computation. To address this issue, BIFE introduces a 
\emph{semantic sparse KV cache} that selectively stores informative tokens from past chunks 
\cite{zhang2025bife}.

Given a chunk $c$, the diffusion transformer produces query, key, and value tensors 
$Q,K,V$. The attention scores are computed as

\begin{equation}
A = \mathrm{Softmax}\!\left(\frac{QK^\top}{\sqrt{d}}\right).
\end{equation}

Token importance is obtained by aggregating attention scores across heads, producing an 
importance vector $m$. Only the top-$k$ tokens with highest importance are retained,

\begin{equation}
I_{\text{keep}} = \text{TopK}(m),
\end{equation}

which forms the sparse cache

\begin{equation}
(K^{\text{sparse}}, V^{\text{sparse}}).
\end{equation}

This token selection mechanism is related to dynamic token sparsification approaches used 
for efficient vision transformers and multimodal models \cite{he2025zipvl}.

During generation, sparse KV entries from previous chunks are stored in a global memory bank. 
Relevant context is retrieved using semantic similarity between prompt embeddings. 
Let $E_c$ denote the embedding of the current prompt and $E_i$ denote the embedding of 
a past chunk $i$. Their similarity is computed as

\begin{equation}
s_i = \cos(E_c, E_i).
\end{equation}

The top-$l$ most similar chunks are selected, and their sparse KV states are combined with 
the most recent chunks to construct the final context

\begin{equation}
(K^*, V^*) = \mathrm{Concat}\big((K_{c-1},V_{c-1}),
(K^{\text{sparse}}_{i_1},V^{\text{sparse}}_{i_1}),\dots\big).
\end{equation}

This semantic sparse KV cache significantly reduces memory consumption while improving 
long-range coherence by retrieving semantically relevant historical context. However, 
the sparse KV entries from all previous chunks must still be stored, causing the cache 
size to grow with sequence length. This limitation motivates the hierarchical KV 
management strategy proposed in WorldAttention.

\section{Metrics}
\label{sec:metric}

For VBench-Long~\cite{huang2024vbench}, we follow prior minute-long video generation works~\cite{guo2025long,cai2025mixture}, we additionally adopt five complementary metrics from VBench~\cite{huang2024vbench} to comprehensively evaluate long-horizon generation quality. These metrics cover both visual fidelity and temporal consistency: (1) \textit{Imaging Quality}, which measures frame-level technical fidelity by quantifying distortions such as over-exposure, noise, and blur; (2) \textit{Motion Smoothness}, which evaluates the continuity and physical plausibility of frame-to-frame transitions; (3) \textit{Aesthetic Quality}, which assesses visual appeal, including composition, color harmony, and photorealism; (4) \textit{Background Consistency}, which measures the stability of the scene background over time; and (5) \textit{Subject Consistency}, which evaluates whether a subject’s appearance remains temporally coherent throughout the video. For InterVBench~\cite{zhang2025bife,team2025inferix}, we follow the metric termed \textit{Video Drift Error (VDE)} to quantify quality changes over time. VDE divides a long video into multiple segments and measures temporal drift for specific attributes, where lower scores indicate better stability. Based on this formulation, we design five drift-aware metrics tailored for long-horizon generation: (1) \textit{VDE Clarity}, which measures temporal drift in image sharpness, penalizing progressive blur; (2) \textit{VDE Motion}, which evaluates drift in motion smoothness, capturing jitter or freezing artifacts; (3) \textit{VDE Aesthetic}, which measures degradation in visual appeal over time; (4) \textit{VDE Background}, which quantifies instability or flickering in scene background; and (5) \textit{VDE Subject}, which tracks identity drift to assess whether the subject remains consistently recognizable throughout the sequence.

\section{Limitation and Future Work}
\label{sec:limitations}

Although WorldAttention demonstrates strong performance for text-conditioned interactive video generation, the current framework primarily focuses on textual prompts as the control signal. This design limits the ability to model richer forms of interaction and control that are common in real-world scenarios. Future work can extend the framework to incorporate additional conditioning modalities, such as action signals, control trajectories, or embodied interaction cues. Integrating action-conditioned generation would enable the model to function as a video world model, supporting controllable environment simulation and decision-driven video prediction.

\section{Kernel Customization}
\label{sec:ko}

To realize the efficiency benefits of Hybrid Sparse Attention (HSA) and Hierarchical KV cache (HKV), we integrate attention kernels customized from VSA\cite{zhang2025vsa} with an optimized KV memory layout to minimize data movement and maximize GPU utilization.

For \textit{HSA}, sparse attention is implemented with two kernels corresponding to the global linear branch and the local sparse branch. The sparse branch attends to token blocks selected by head-adaptive sparsity, while the linear branch captures long-range dependencies with linear complexity. 

For \textit{HKV}, cached KV states are organized into semantically grouped pages in a hierarchical layout. During decoding, a lightweight retrieval stage selects relevant KV pages, which are mapped to contiguous memory before attention computation, enabling consolidated access and efficient kernel execution.

\section{Results on InterVBench}
\label{sec:intervbench}

\begin{table*}[h]
  \centering
  \caption{\textbf{Comparison of different methods on InterVBench.} We report InterVBench results on five VDE metrics and five complementary metrics from VBench \cite{huang2024vbench}.}
  \label{tab:intervbench}
  \setlength{\tabcolsep}{6pt}
  \renewcommand{\arraystretch}{1.2}
  \resizebox{0.8\linewidth}{!}{
    \begin{tabular}{l|c|c|c|c|c}
      \toprule
      Method &
      \makecell{VDE \\ Subject} $\downarrow$ &
      \makecell{VDE \\ Background} $\downarrow$ &
      \makecell{VDE \\ Motion} $\downarrow$ &
      \makecell{VDE \\ Aesthetic} $\downarrow$ &
      \makecell{VDE \\ Clarity} $\downarrow$ \\
      \midrule
      MAGI-1 & 0.3090 & 0.5000 & 0.0243 & 3.8286 & 2.7225\\
      Self Forcing & 0.3716 & 1.6108 & 0.1549 & 3.4683 & 3.0798\\
      PAVDM & 1.8292 & 0.9323 & 0.0461 & 2.8957 & 1.9503\\
      FramePack & 4.3984 & 5.9421 & 0.0387 & 1.4751 & 4.2513 \\
      SkyReels-V2-DF-1.3B & 0.1085 & 0.3179 & 0.0195 & 1.2083 & 0.9365 \\
      LongLive & 0.0825 & 0.3016 & 0.0104 & 1.1452 & 0.8529 \\
      \midrule
      BIFE & 0.0844 & 0.2945 & 0.0119 & 0.9618 & 0.7551\\
      \rowcolor{Periwinkle!30}\textbf{WorldAttention (Ours)} & \textbf{0.0792} & \textbf{0.2850} & \textbf{0.0095} & \textbf{0.9462} & \textbf{0.7390}\\
      \midrule
      Method &
      \makecell{Subject \\ Consistency} $\uparrow$ &
      \makecell{Background \\ Consistency} $\uparrow$ &
      \makecell{Motion \\ Smoothness} $\uparrow$ &
      \makecell{Aesthetic \\ Quality} $\uparrow$ &
      \makecell{Image \\ Quality} $\uparrow$ \\
      \midrule
      MAGI-1 & 0.8992 & 0.9078 & 0.9947 & 0.6508 & 0.6662\\
      Self Forcing & 0.8481 & 0.8203 & 0.9947 & 0.6283 & 0.6805\\
      PAVDM & 0.8640 & 0.8924 & 0.9926 & 0.5267 & 0.6567\\
      FramePack & 0.9001 & 0.8791 & 0.9949 & 0.6043 & \textbf{0.6972} \\
      SkyReels-V2-DF-1.3B & 0.9418 & 0.9579 & 0.9931 & 0.6035 & 0.6835 \\
      LongLive & 0.9610 & 0.9552 & 0.9968 & \textbf{0.6595} & 0.6960 \\
      \midrule
      BIFE & 0.9597 & 0.9588 & 0.9956 & 0.6047 & 0.6852\\
      \rowcolor{Periwinkle!30}\textbf{WorldAttention (Ours)} & \textbf{0.9668} & \textbf{0.9614} & \textbf{0.9972} & 0.6397 & 0.6965\\
      \bottomrule
    \end{tabular}
  }

\end{table*}

\paragraph{\upshape\textbf{InterVBench.}}
We further evaluate WorldAttention under the chunk-level long-video generation setting of InterVBench, which contains 200 test videos longer than 50 seconds with fine-grained annotations. Following the standard evaluation protocol, we report five VDE-based drift metrics together with five complementary VBench metrics to assess both temporal robustness and perceptual quality. As shown in Table~\ref{tab:intervbench}, WorldAttention achieves state-of-the-art performance across the majority of metrics, demonstrating improved long-horizon stability and overall generation quality compared to prior methods.

\section{Sparse Branch}
\label{sec:hsa_details}

Unlike fixed Top-$K$ block selection, HSA introduces \emph{head-adaptive sparsity}. Each attention head independently determines its sparsity level based on the observed distribution of its attention scores. Specifically, for each head, we sort blocks by attention score in descending order and select the minimal set whose cumulative attention exceeds a predefined threshold. This yields a spectrum of sparsity patterns across heads: some heads become highly sparse and focus on local interactions, others are moderately sparse and capture structured dependencies, while a few retain low sparsity to preserve broader context. Such head‑level adaptation offers finer granularity than layer‑level or global sparsity schemes.

Fixed Top-$K$ block selection is replaced by a \emph{cumulative attention score} criterion \emph{per head}.
For the $i$-th query block and $h$-th head, the block attention weights are denoted by $\mathbf{p}^{(h)}_{i,:}\in\mathbb{R}^{N}$, representing the $i$-th row of $\mathbf{P}^{(h)}$.
Let $\sigma^{(h)}_i$ be a permutation that sorts blocks by descending attention weight:
\[
\mathbf{p}^{(h)}_{i,\sigma^{(h)}_i(1)} \ge \mathbf{p}^{(h)}_{i,\sigma^{(h)}_i(2)} \ge \cdots \ge \mathbf{p}^{(h)}_{i,\sigma^{(h)}_i(N)}.
\]
Given a head-specific coverage threshold $\tau_h\in(0,1]$, we choose the \emph{minimal} number of blocks
\[
K^{(h)}_i
=
\min\Big\{k:\;\sum_{m=1}^{k}\mathbf{p}^{(h)}_{i,\sigma^{(h)}_i(m)} \ge \tau_h\Big\}.
\]
The selected key-block set for $(i,h)$ is then
\[
\mathcal{S}^{(h)}(i)
=
\Big\{\sigma^{(h)}_i(1),\dots,\sigma^{(h)}_i\!\big(K^{(h)}_i\big)\Big\}.
\]
Finally, the sparse branch uses
\[
\mathcal{S}^{(h)} \;=\; \bigcup_{i=1}^{N}\{(i,j): j\in \mathcal{S}^{(h)}(i)\},
\]
which directly specifies the set of blocks executed by the block-sparse attention kernel.

To determine the head-specific threshold $\tau_h$, we quantify the sparsity of the attention distribution using the Gini coefficient. Using the block attention weights $\mathbf{p}^{(h)}_{i,:}$ and  descending permutation $\sigma^{(h)}_i$, the Gini coefficient $\mathcal{G}^{(h)}$ for the $h$-th head is calculated across all query blocks:
\begin{equation}
\mathcal{G}^{(h)} = \mathbb{E}_i \left[ \frac{1}{N} \sum_{m=1}^{N} (N - 2m + 1) \cdot \mathbf{p}^{(h)}_{i,\sigma^{(h)}_i(m)} \right].
\end{equation}
Here, $\mathcal{G}^{(h)}$ serves as a scalar proxy of sparsity, taking higher values for peaked distributions and lower values for uniform ones. We then map this metric to the coverage threshold $\tau_h$ via:
\begin{equation}
\tau_h = \tau_{\min} + \mathcal{G}^{(h)} \cdot (\tau_{\max} - \tau_{\min}),
\end{equation}
where $\tau_{\min}$ and $\tau_{\max}$ are hyperparameters that bound the cumulative mass.

This adaptive scaling dynamically determines the number of selected blocks per head: higher $\tau_h$ selects more blocks for high-sparsity heads, while lower $\tau_h$ limits block selection for low-sparsity heads, maintaining computational efficiency.

\section{Linear Branch}
\label{sec:linear}

\begin{figure}[t]
    \centering
    \includegraphics[width=\linewidth]{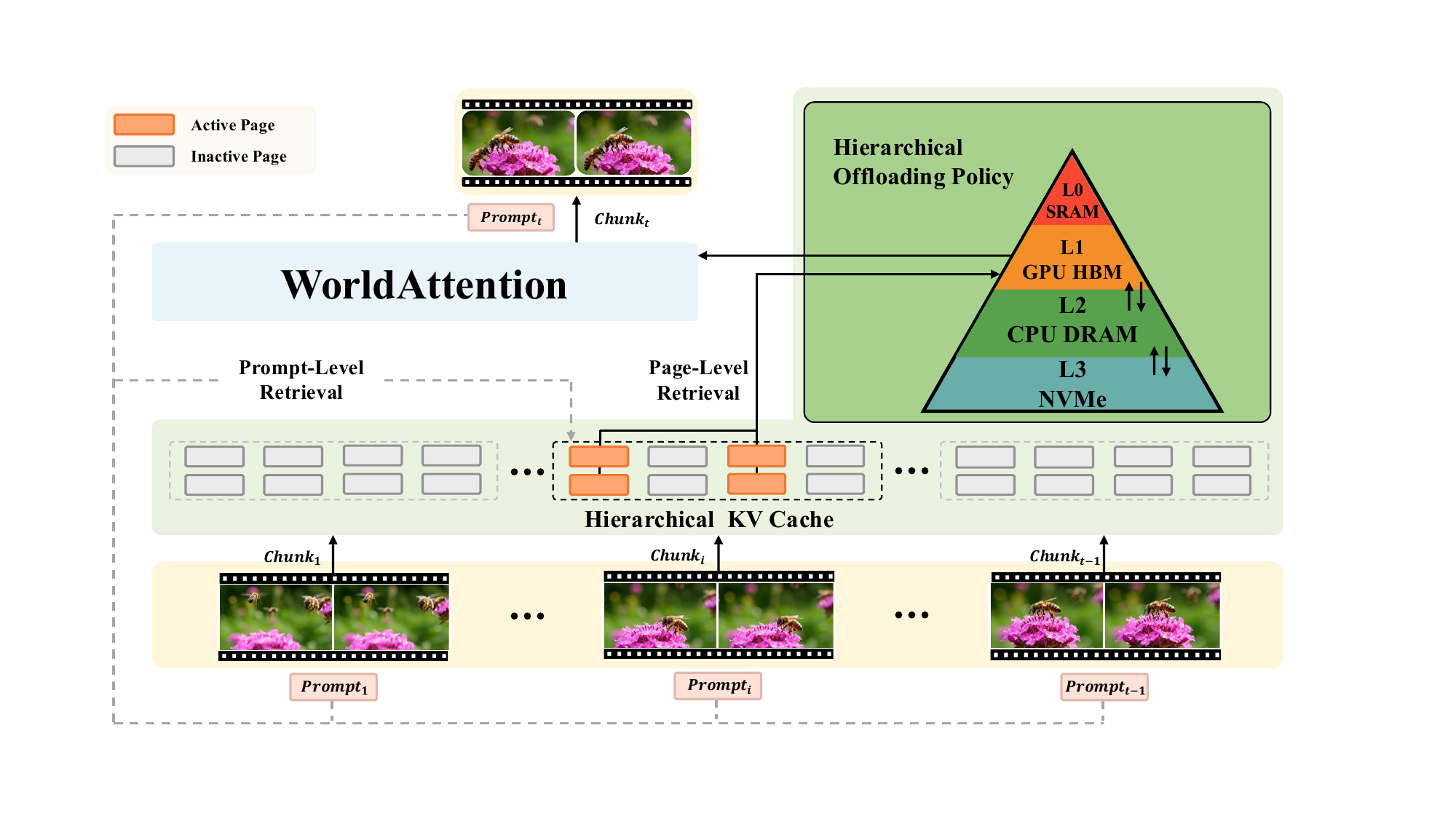}
    \caption{Overview of the WorldAttention architecture featuring a three-tier hierarchical KV cache and a two-stage (prompt-level and page-level) retrieval mechanism for efficient long-video generation.}
    \label{fig:pipeline}
    \vspace{-1em}
\end{figure}

Many efficient architectures, such as SLA \cite{zhang2025sla}, adopt linear attention to reduce computational complexity. Their core mechanism relies on continuously compressing unbounded historical information into a fixed-size hidden state matrix. However, recent studies \cite{sse2025, xiao2026ram} show that this forced compression introduces a fundamental capacity bottleneck, inevitably leading to memory superposition and attention dilution. To overcome this limitation, we employ Linformer with an explicit structured spatial downsampling paradigm. To handle dynamically growing sequences with fixed-size projection matrices $\mathbf{E}_K, \mathbf{E}_V \in \mathbb{R}^{L \times r}$, we partition the retrieved KV tokens into fixed-length segments of size $L$. Each segment is projected independently and then concatenated to form the final compressed context. By performing low-rank projection and concatenation on the historical KV cache at the chunk/page level, this mechanism not only aligns with the low-rank property of attention matrices in long sequences \cite{wang2020linformer}, but more importantly, preserves exact spatiotemporal physical anchors and relative positional topologies for long-range information. 

This theoretical analysis is empirically validated in our experiments. As illustrated in Fig.~\ref{fig:attn_analysis}(a) and (b), standard linear attention exhibits severe feature oversmoothing over long sequences, failing to maintain the attention peaks present in full attention at the initial frame and within local contexts. This shows that implicit state accumulation fails to accurately localize and extract key visual cues in long-video interactions. In contrast, Linformer accurately recovers the long-range attention mass highly consistent with full attention, capturing the peak distributions of key contexts both globally and in the case of the last chunk. This precise restoration directly translates into a significant advantage in representational capacity: as shown in \textbf{Fig.~\ref{fig:attn_analysis}(c)}, Linformer consistently outperforms standard linear attention by a large margin in terms of output cosine similarity relative to the full attention baseline. This fully demonstrates that in long-range interactive scenarios, Linformer can achieve high-fidelity extraction of fine-grained visual features with near-linear complexity.

\clearpage
\section{Visualization}
\label{sec:vis}

\begin{figure*}[h]
\centering
\includegraphics[width=\linewidth]{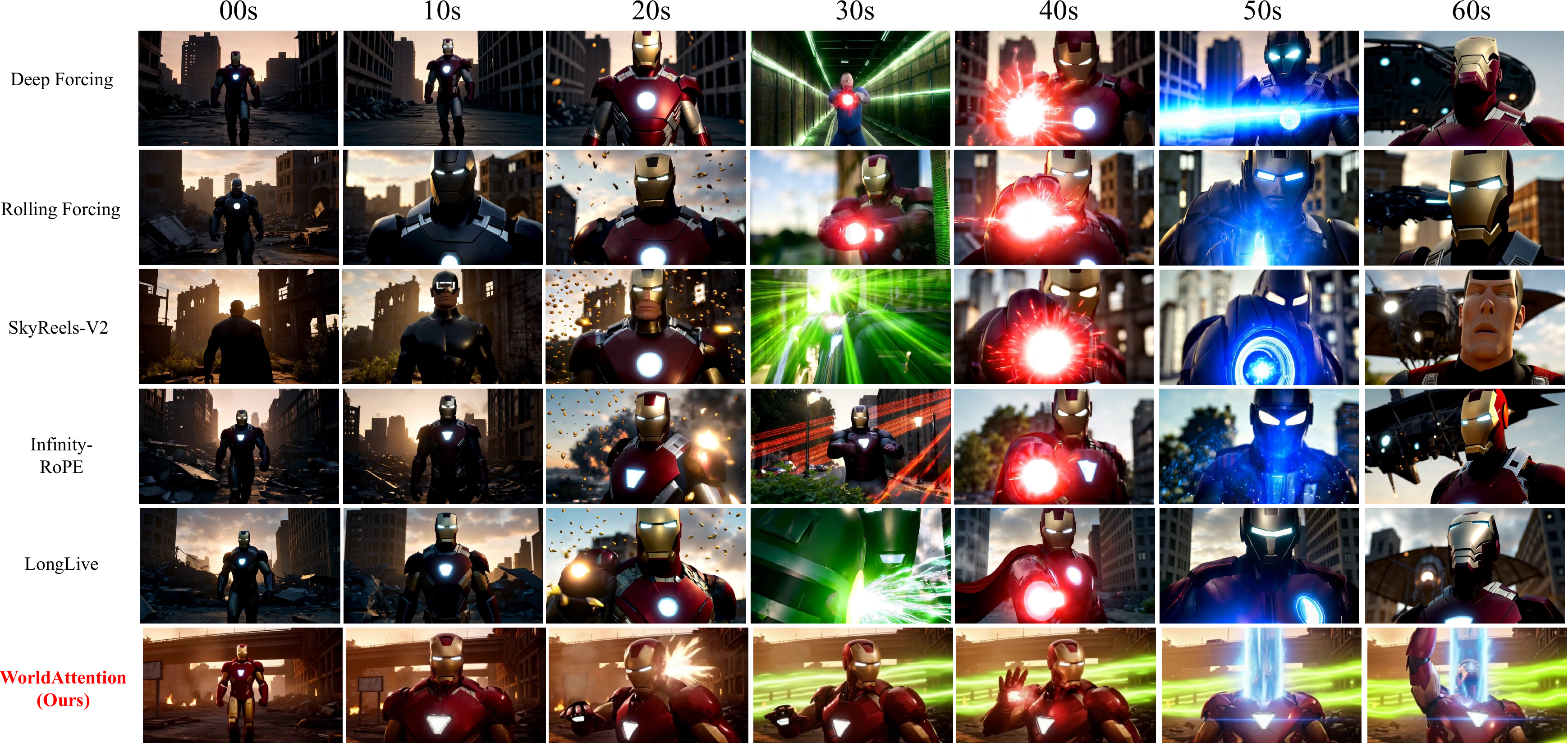}
\caption{An interactive long-video generation example.}
\label{fig:vis_1}
\end{figure*}

\begin{figure*}[t]
\centering
\includegraphics[width=\linewidth]{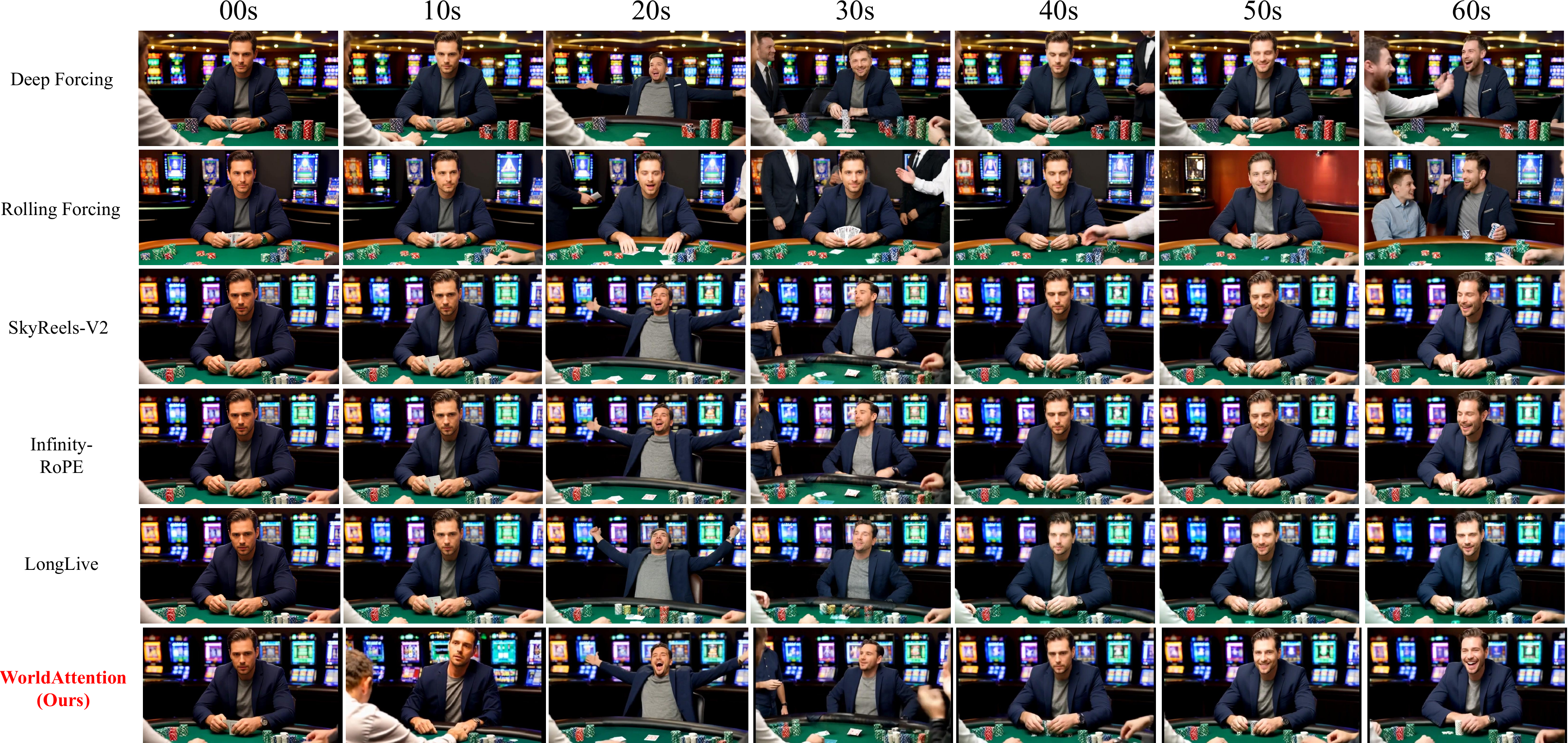}
\caption{An interactive long-video generation example.}
\label{fig:vis_2}
\end{figure*}

\begin{figure*}[t]
\centering
\includegraphics[width=\linewidth]{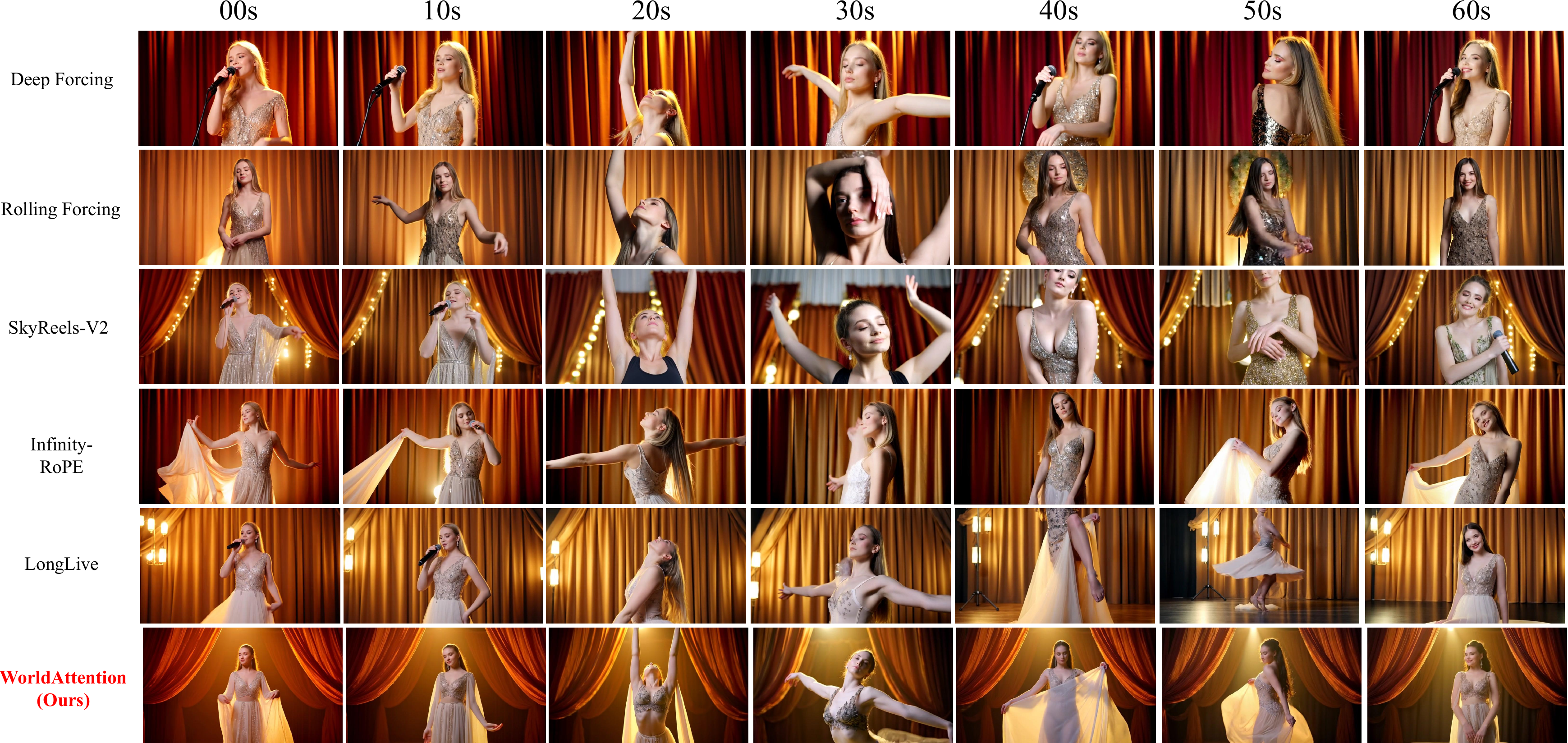}
\caption{An interactive long-video generation example.}
\label{fig:vis_3}
\end{figure*}

\begin{figure*}[t]
\centering
\includegraphics[width=\linewidth]{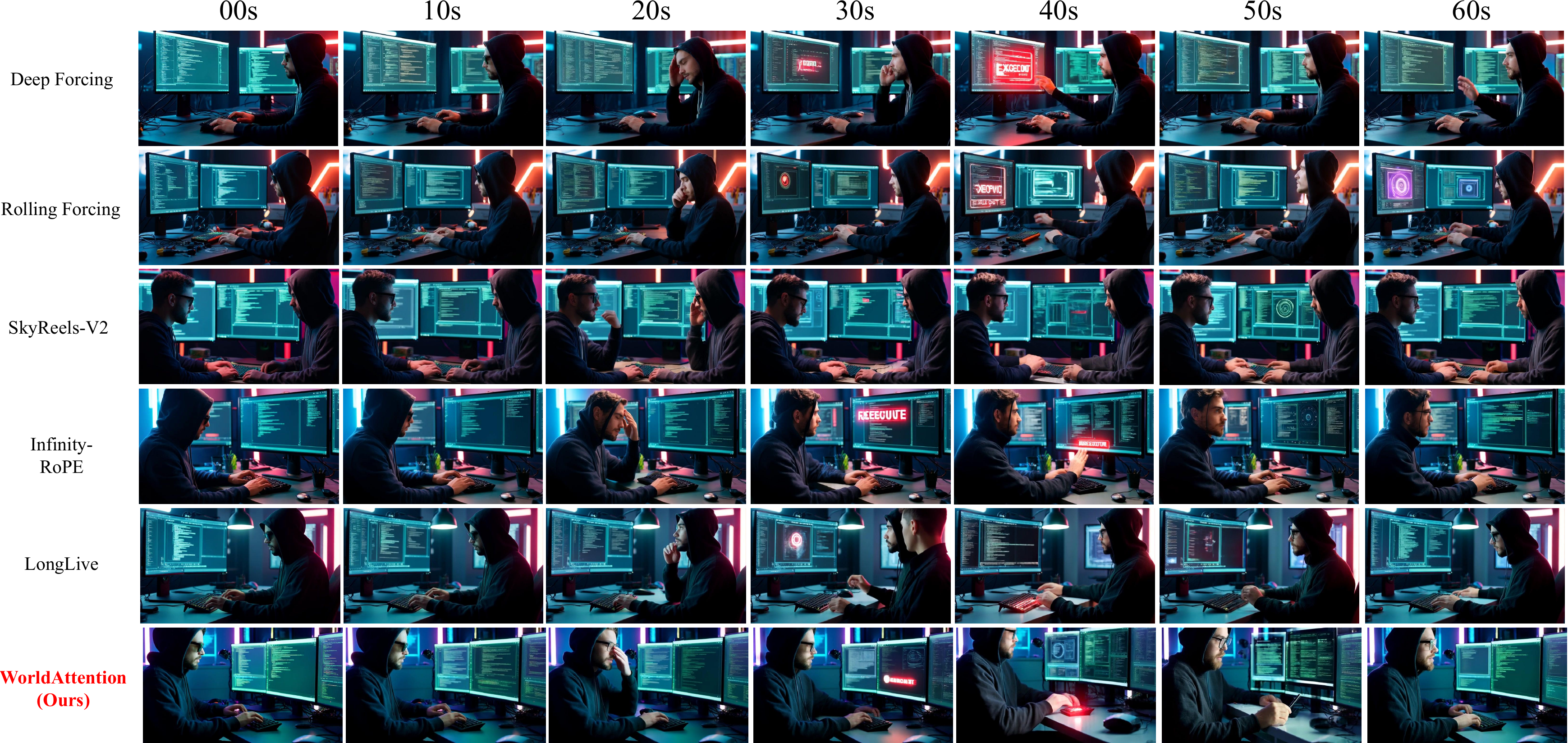}
\caption{An interactive long-video generation example.}
\label{fig:vis_4}
\end{figure*}

\begin{figure*}[t]
\centering
\includegraphics[width=\linewidth]{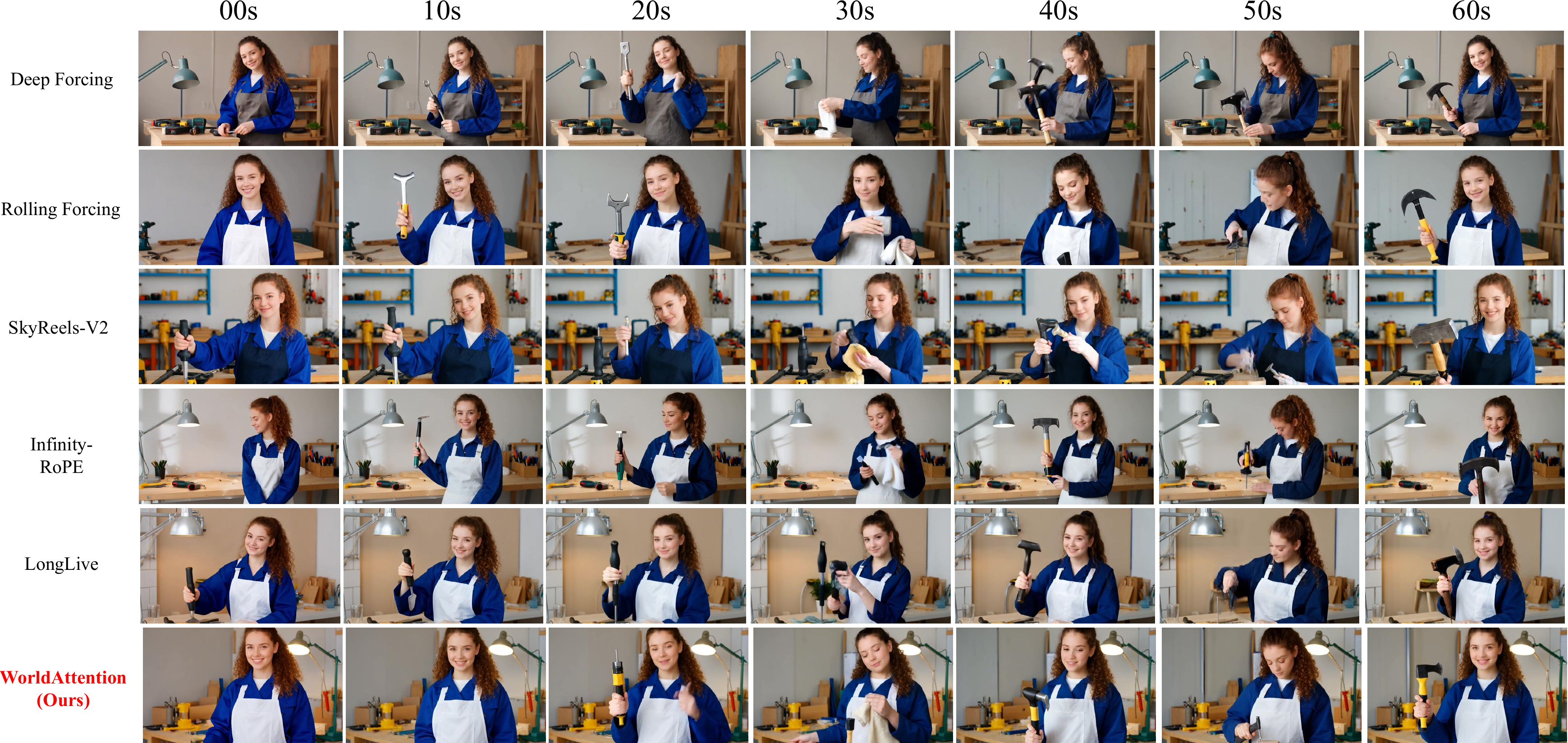}
\caption{An interactive long-video generation example.}
\label{fig:vis_5}
\end{figure*}

\begin{figure*}[t]
\centering
\includegraphics[width=\linewidth]{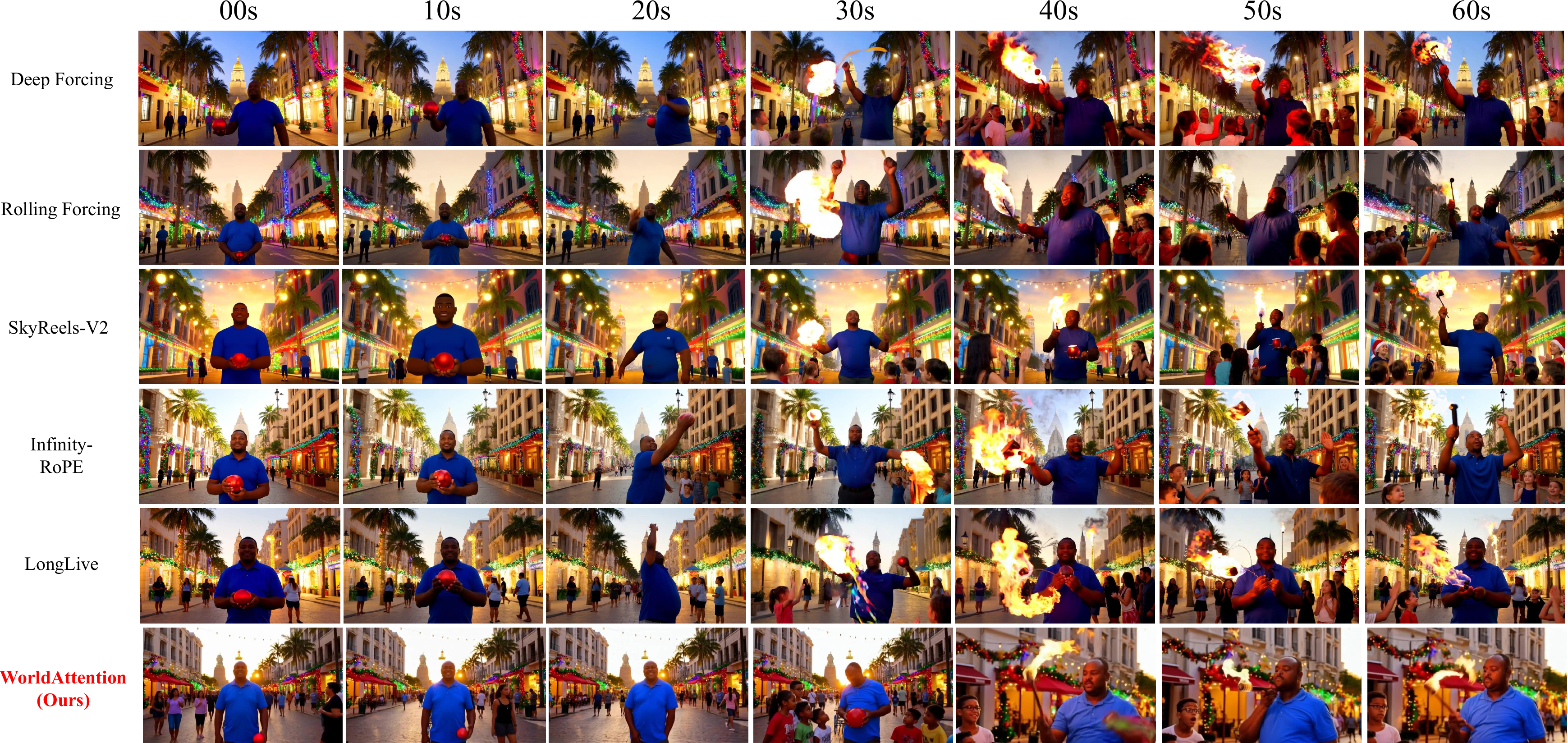}
\caption{An interactive long-video generation example.}
\label{fig:vis_6}
\end{figure*}

Figures~\ref{fig:vis_1}, \ref{fig:vis_2}, \ref{fig:vis_3}, \ref{fig:vis_4}, \ref{fig:vis_5}, and \ref{fig:vis_6} present qualitative comparisons on an interactive long-video generation example. 
Compared with existing approaches, WorldAttention maintains stronger subject consistency, smoother motion transitions, and more stable scene structure across the entire sequence. 

\begin{center}
\begin{tcolorbox}[promptbox, fontupper=\small, before skip=0pt, after skip=0pt]

"captions": [

"Iron Man walks through a war-torn, crumbling urban ruin at dusk, then stops and stands still. Wide shot to medium close-up (00s - 09s).",

"A hail of bullets tears in. Iron Man raises his arm, rounds and shells streak past. Wide shot to medium close-up (10s - 19s).",

"Emerald-green beams sweep down the street. Wide shot to medium close-up (20s - 29s).",

"Iron Man aims and fires tight, pulsed red repulsor blasts from his palm. Wide shot to medium close-up (30s - 39s).",

"The chest arc reactor unleashes a colossal blue beam. Wide shot to medium close-up (40s - 49s).",

"A dark alien craft enters from the left and cruises across. Iron Man raises his arm again while looking up at the alien craft. Wide shot to medium close-up (50s - 59s)."]

\end{tcolorbox}
\end{center}

\newpage

\begin{center}
\begin{tcolorbox}[promptbox, fontupper=\small, before skip=0pt, after skip=0pt]

"captions": [

"A realistic video of a Texas Hold'em poker game in a casino. A late-30s male player with short dark hair and light stubble, wearing a navy blazer, charcoal tee, dark jeans, and a stainless-steel watch, sits at a well-lit poker table holding his cards close. Colorful chips are spread across the felt, the dealer deals steadily, and slot machines glow in the background. Wide shot to medium close-up (00s - 09s).",

"At the same poker table, the same player throws his cards onto the felt and leans back in his chair, arms opening as the action ends. The dealer continues dealing and the table remains busy with chips and players. Wide shot to medium close-up (10s - 19s).",

"Moments later, the same player reveals the winning hand on the table and stays leaned back confidently. A nearby patron claps while the dealer moves the game forward. Casino tables and slot machines fill the background. Wide shot to medium close-up (20s - 29s).",

"After the result, the same player sits forward and stacks his chips neatly, aligning the piles with steady hand movements. The dealer begins the next round as the casino lights flicker softly behind. Wide shot to medium close-up (30s - 39s).",

"The same player pauses and looks over his organized chips, showing a small, satisfied smile. The table stays active and well lit. Wide shot to medium close-up (40s - 49s).",

"To finish the sequence, the same player turns and exchanges a celebratory high-five with a nearby patron. Brief cheers rise as the dealer keeps dealing and the casino continues around them. Wide shot to medium close-up (50s - 59s)."]

\end{tcolorbox}
\end{center}

\newpage

\begin{center}
\begin{tcolorbox}[promptbox, fontupper=\small, before skip=0pt, after skip=0pt]

"captions": [

"A young white woman in a flowing, sequined gown performs a graceful song-and-dance on a warmly lit stage with swaying curtains behind her. Medium close-up (00s - 09s).",

"The same stage setup continues as the dancer leaps with her arms extended overhead, and the camera follows her fluid rise under the warm light. Medium close-up (10s - 19s).",

"The same dancer lands softly, opening her arms as if embracing the stage, while her movement remains smooth and poised in front of the swaying curtains. Medium close-up (20s - 29s).",

"The same warm stage remains in view as a spotlight slides down to her feet, revealing crisp, rhythmic steps beneath the flowing sequined gown. Medium close-up (30s - 39s).",

"The same dancer twirls gracefully; her skirt catches the stage light as the spotlight beam tracks her turning feet. Medium close-up (40s - 49s).",

"The same young woman settles under the warm glow and offers a radiant smile, full of poise and passion, with the curtains swaying softly behind her. Medium close-up (50s - 59s)."]

\end{tcolorbox}
\end{center}

\newpage

\begin{center}
\begin{tcolorbox}[promptbox, fontupper=\small, before skip=0pt, after skip=0pt]

"captions": [

"A hacker wearing sunglasses and a hoodie types at flickering monitors inside a cyberpunk workshop. Neon shadows rake across a cluttered workbench as streams of code slide across the screens. Medium close-up (00s - 09s).",

"The same cyberpunk workshop setup remains in view as the hacker pauses, leans back, and rubs their eyes while the monitors keep strobing with code and neon light flickers across the cluttered bench. Medium close-up (10s - 19s).",

"The same hacker stays focused at the flickering monitors as a small red alert begins blinking on one screen, drawing immediate attention while code continues streaming through the neon-lit workshop. Medium close-up (20s - 29s).",

"The same cyberpunk workshop scene continues as the hacker's hand hesitates over a red EXECUTE button, then presses it decisively while the monitors flicker and neon shadows cut across the cluttered workbench. Medium close-up (30s - 39s).",

"The same medium close-up remains in the cyberpunk workshop as a nearby device's LED flashes confirmation, and the hacker glances toward it while the monitors continue strobing and code streams by. Medium close-up (40s - 49s).",

"The same hacker remains in the neon-lit workshop as a second device erupts in rapid beeps. They pivot toward it, calm and intent, while flickering monitors and scrolling code light the cluttered bench. Medium close-up (50s - 59s)."]

\end{tcolorbox}
\end{center}

\newpage

\begin{center}
\begin{tcolorbox}[promptbox, fontupper=\small, before skip=0pt, after skip=0pt]

"captions": [

"A young white woman in a blue work coat and white apron, with a curly brown ponytail, lifts a tool from a tidy, well-lit workbench and smiles to the camera in greeting. Medium close-up, static shot (00s - 09s).",

"At the same tidy workbench, the same young woman in the blue work coat and white apron tightens a screw with the tool, then shows the result to the camera with a satisfied nod. Medium close-up, static shot (10s - 19s).",

"The same woman wipes her hands on a rag and gestures proudly toward the finished piece on the neat workbench, still framed in the same bright workshop setting. Medium close-up, static shot (20s - 29s).",

"Still smiling, the same young woman picks up a small hammer from the tidy workbench to demonstrate the next step, her curly brown ponytail visible above the white apron. Medium close-up, static shot (30s - 39s).",

"The same woman gently taps a nail into wood with steady, careful strikes. Her curly brown ponytail sways slightly as she works at the clean, well-lit workbench. Medium close-up, static shot (40s - 49s).",

"Leaning closer to the camera with bright eyes, the same young woman offers a proud, friendly smile while the small hammer rests in her hand beside the finished piece. Medium close-up, static shot (50s - 59s)."]

\end{tcolorbox}
\end{center}

\newpage

\begin{center}
\begin{tcolorbox}[promptbox, fontupper=\small, before skip=0pt, after skip=0pt]

"captions": [

"A realistic vibrant Christmas street scene in Rio de Janeiro with colorful holiday lights, festive decorations, palm trees, tall buildings, and locals dressed in summer clothes. At the center, a slightly overweight Black man in a blue shirt holds a small red ball. Christ the Redeemer peeks in the distance, and a warm golden glow fills the scene. Wide shot to medium close-up (00s - 09s).",

"The same vibrant Christmas street in Rio de Janeiro remains full of colorful lights, festive decor, palm trees, tall buildings, and locals in summer wear. The same slightly overweight Black man in a blue shirt tosses the small red ball into the air, drawing a small crowd of kids and adults around him. Christ the Redeemer is still visible in the distance under the warm golden glow. Wide shot to medium close-up (10s - 19s).",

"The same performer in the blue shirt switches from the red ball to long ribbon streamers on the lively Christmas street. He twirls colorful ribbons through the warm golden air as kids and adults gather closer, cheering amid holiday lights, festive decorations, palm trees, and tall buildings, with Christ the Redeemer still peeking in the distance. Wide shot to medium close-up (20s - 29s).",

"The same slightly overweight Black man in the blue shirt now performs with a flaming torch on the vibrant Christmas street. The crowd's excitement surges as the torch flame flickers against colorful holiday lights and festive decor, with palm trees, tall buildings, summer-dressed locals, and Christ the Redeemer in the distance. Wide shot to medium close-up (30s - 39s).",

"The same street performer balances the flaming torch carefully on his chin while standing at the center of the Christmas street. Children press closer with wide-eyed wonder as adults cheer behind them, surrounded by colorful lights, festive decorations, palm trees, tall buildings, and the warm golden glow of Rio with Christ the Redeemer in the distance. Wide shot to medium close-up (40s - 49s).",

"The same slightly overweight Black man in the blue shirt twirls the flaming torch between his fingers as applause rolls through the audience. Kids and adults cheer on the vibrant Christmas street, framed by colorful holiday lights, festive decor, palm trees, tall buildings, summer clothes, a warm golden glow, and Christ the Redeemer peeking in the distance. Wide shot to medium close-up (50s - 59s)."]

\end{tcolorbox}
\end{center}

\end{document}